\documentclass[11pt]{article}
\usepackage{hyperref}
\usepackage{url}
\usepackage{amsmath}
\usepackage{bm}
\usepackage{amsfonts}
\usepackage{algorithm}
\usepackage{algorithmic}
\usepackage{amsthm,paralist}
\usepackage{enumitem,graphicx}
\usepackage{color}
\usepackage{graphics, subfigure}

\usepackage[scriptsize,bf]{caption}
\usepackage{amssymb}

\usepackage[numbers,square]{natbib}

\newcommand{\beq}{\begin{equation}}
\newcommand{\eeq}{\end{equation}}

\newtheorem{theorem}{Theorem}[section]

\newtheorem{claim}{Claim}[section]

\newtheorem{assumption}{Assumption}[section]

\newif\ifarxiv

 \arxivtrue 

 \newif\ifSPmag
 
 \SPmagfalse

\newif\ifjournal

 \journalfalse

\newif\ifnotsure   

\notsurefalse 

\title{  Optimization  for deep learning:  theory and algorithms }  

\author{Ruoyu Sun \thanks{
	 Department of Industrial and Enterprise Systems Engineering (ISE), and
	 affiliated to Coordinated Science Laboratory and Department of ECE,
	  University of Illinois at Urbana-Champaign, Urbana, IL.
 Email: ruoyus@illinois.edu.  }
	\date{\today} 
}

\begin{document}

\maketitle
\vspace{-0.5cm}
\begin{abstract}

\ifarxiv 
When and why can a neural network be successfully trained?
This article provides an overview of optimization algorithms and theory  for training  neural networks. First, we  discuss the issue of gradient explosion/vanishing and the more general issue of undesirable spectrum, and then discuss  practical solutions including careful initialization and normalization methods. Second, we review generic optimization methods used in training neural networks, such as SGD, adaptive gradient methods and 
distributed methods, and existing theoretical results for these algorithms. Third, we review existing research  on the global issues of neural network training, including results on bad local minima, mode connectivity, lottery ticket hypothesis and infinite-width analysis. 

\fi

\end{abstract}

\vspace{-0.3cm}
\section{Introduction}
\vspace{-0.3cm}

\ifarxiv 
A major theme of  this article is to understand the practical components
for successfully training neural networks, and the possible factors that cause the failure of training. 
Imagine you were in year 1980 trying to solve an image classification
problem  using neural networks.   
If you wanted to train a neural network from scratch, 
it is very likely that your first few attempts would have failed to return reasonable results. What are the essential changes to make the algorithm work?
In a high-level, you need three things (besides powerful hardware):
a proper neural network, a proper training algorithm, and proper training tricks. 

\begin{itemize}
	\item Proper neural-net. This includes neural architecture and activation functions.
	For neural architecture, you may want to replace a fully connected network by a convolutional network with at least 5 layers and enough neurons.
	For better performance,
	you may want to increase the depth to 20 or even 100, and add skip connections. 
	For activation functions, a good starting point
	is ReLU activation, but using tanh or swish activation is also reasonable.


	\item Training algorithm. 
	 A big choice is to use stochastic versions of gradient descent (SGD)
	 and stick to it.  A well-tuned constant step-size is good enough, while  momentum 
    and adaptive stepsize can provide extra benefits. 

	\item Training tricks. 
Proper initialization is very important for the algorithm to start training.
To train a network with more than 10 layers,  two extra tricks are often needed: 
adding normalization layers and adding skip connections.

\end{itemize}

%

Which of these design choices  are essential?  
Currently we have some understanding of a few design choices, including initialization
strategies, normalization methods, the skip connections,
over-parameterization (large width) and SGD, as shown in Figure \ref{fig0}.
We roughly divide the optimization advantage into three parts:
 controlling Lipschitz constants, faster convergence and better landscape.
There are many other design choices that are hard to understand,
most notably the neural architecture.  
 Anyhow, it seems  impossible to understand every part of this complicated system, and the current understanding can already provide some useful insight.  

\begin{figure}
	\centering
	\includegraphics[width=10cm,height=4cm]{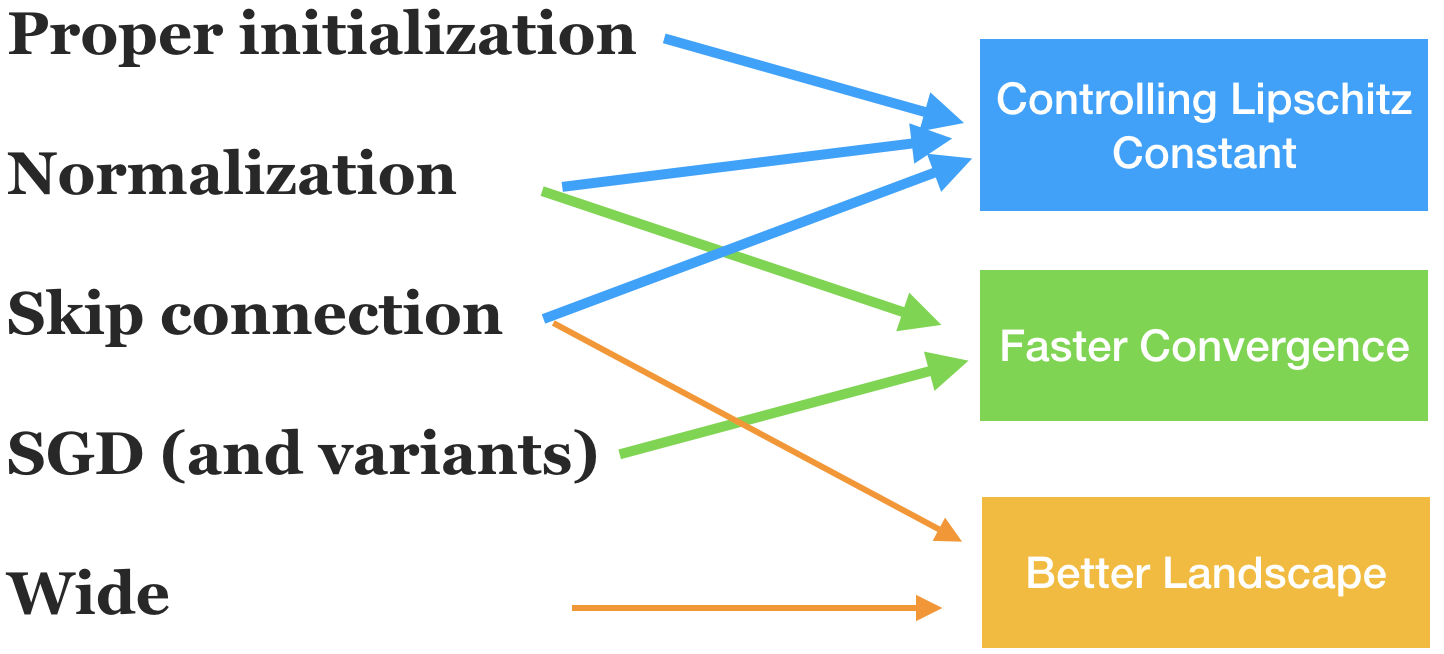}
	\caption{A few major design choices for a successful  training of a neural network
		with theoretical understanding. 
		They have impact on three aspects of algorithm convergence:  make convergence possible, faster convergence and  better global solutions.
		The three aspects are somewhat related, and it is jut a rough classification. 
		Note that there are other important design choices, especially the neural architecture, 
		that is not understood theoretically, and thus omitted in this figure. 
		There are also other benefits such as generalization, which is also omitted. 
	}
	\label{fig0}
\end{figure}

\fi 


To keep the survey simple, we will focus on the supervised learning problem with feedforward neural networks.
We  will  not discuss more complicated formulations such as GANs (generative adversarial networks) and deep reinforcement learning, and do not discuss more complicated architecture such as RNN (recurrent neural network), attention and Capsule. 
In a broader context, theory for supervised learning contains at least  representation, optimization and generalization (see Section \ref{subsec: big picture of decomposition}), and we do not discuss representation and generalization in detail.
One major  goal is to understanding how the \textit{neural-net structure} (the parameterization by concatenation of many variables) affects the design and analysis of optimization algorithms, which can potentially go beyond supervised learning.

This article is written for  researchers who are interested in 
theoretical understanding of optimization for neural networks.
Prior knowledge on optimization methods and basic theory will be very helpful (see, e.g., \cite{bertsekas1997nonlinear,sra2012optimization,bottou2018optimization} for preparation). 
Existing surveys on optimization for deep learning are intended for general machine learning audience, such as Chapter 8 of the  book Goodfellow et al. \cite{goodfellow2016deep}. 
These reviews often do not discuss optimization theoretical aspects in depth.
In contrast, in this article, we emphasize more on the theoretical results while trying to make it accessible for non-theory readers. 
 Simple examples that illustrate the intuition will be provided if possible, 
 and we will not explain the details of the theorems.

 \subsection{Big picture: decomposition of theory}\label{subsec: big picture of decomposition}


A  useful and popular meta-method to develop theory is decomposition. 
We first briefly review the role of optimization in machine learning, and then discuss
 how to decompose the theory of optimization for deep learning. 

\textbf{Representation, optimization and generalization.}
The goal of supervised learning is to find a function that approximates
the underlying function based on observed samples. 
The first step is to find a rich family of functions (such as neural networks) that can represent the desirable function.
The second step is to identify the parameter of the function by minimizing a certain loss function.
The third step is to use the function found in the second step to make predictions
on unseen test data, and the resulting error is called test error.
The test error can be decomposed into representation error, optimization error and generalization error, corresponding to the error caused by each of the three steps. 

In machine learning, the three subjects representation, optimization and generalization
are often studied separately. 
For instance, when studying representation power of a certain family of functions,
 we often do not care whether the optimization problem can be solved well. 
 When studying the generalization error, we often assume that the global optima
 have been found (see \cite{jakubovitz2019generalization} for a survey of generalization).  
 Similarly, when studying optimization properties, we often do not explicitly consider the generalization error (but sometimes we assume the representation error is zero).

 \textbf{Decomposition of optimization issues.}
Optimization issues of deep learning are rather complicated, and further decomposition
is needed. The development of optimization can be divided into three steps. 
The first step is to make the algorithm start running and
converge to a reasonable solution such as a stationary point. 
The second step is to make the algorithm converge as fast as possible.
The third step is to ensure the algorithm converge to a solution with 
a low objective value (e.g. global minima).
There is an extra step of achieving good test accuracy, but 
this is beyond the scope of optimization. 
In short, we divide the optimization issues into three parts:
 convergence, convergence speed and global quality. 

$$
 \text{Optimization issues} 
  \begin{cases}
  \text{ Local issues}  \begin{cases}
  \text{ Convergence issue: gradient explosion/vanishing } \\
   \text{ Convergence speed issue  }
   \end{cases}   \\
   \text{ Global issues:  bad local minima, plaeaus, etc. }
  \end{cases}
$$
Most works are reviewed in three sections: Section \ref{sec: gradient explosion and vanishing},
Section \ref{sec: general algorithms} and Section \ref{sec: global optimization}. 
Roughly speaking, each section is mainly motivated by
 one of the three parts of optimization theory.
However,  this partition is not precise as the boundaries between the three parts  are blurred. 
For instance, some techniques discussed in Section \ref{sec: gradient explosion and vanishing}
can also improve the convergence rate, 
and some results in Section \ref{sec: global optimization} address
 the convergence issue as well as global issues. 
Another reason of the partition is that they represent three rather separate subareas of neural network optimization, and are developed somewhat independently.

\subsection{Outline}

The structure of the article is as follows. 
 In Section \ref{sec: formulation}, we present the formulation of a typical neural network optimization problem for supervised learning.
In Section \ref{sec: GD}, we present back propagation (BP) and analyze the difficulty of applying classical convergence analysis to gradient descent for neural networks. 
In Section \ref{sec: gradient explosion and vanishing}, we discuss neural-net specific tricks for training a neural network, and some underlying theory. 
These are neural-network dependent methods, that open the black box of neural networks. 
In particular, we discuss a major challenge called gradient explosion/vanishing
 and a more general challenge of controlling spectrum, 
 and review main solutions such as careful initialization and normalization methods.
In Section \ref{sec: general algorithms}, we discuss generic algorithm design which treats neural networks as generic non-convex optimization problems. 
In particular,  we review SGD with various learning rate schedules, adaptive gradient methods, 
 large-scale distributed training, second order methods and the existing convergence and iteration complexity results.
In Section  \ref{sec: global optimization}, we review research on global optimization of neural networks, including global landscape, mode connectivity, lottery ticket hypothesis
and infinite-width analysis (e.g. neural tangent kernel).


\section{ Problem Formulation }\label{sec: formulation}

In this section, we present the optimization formulation for a supervised learning problem. 
Suppose we are given data points  $x_i \in \mathbb{R}^{d_x}, y_i \in \mathbb{R}^{ d_y } , i=1, \dots, n$, where $n$ is the number of samples.  The input instance $x_i$ can represent a feature vector of an object, an image, a vector that presents a word, etc. 
The output instance $y_i$ can represent a real-valued vector or scalar such as in
a regression problem, or an integer-valued vector or scalar such as in a classification problem.

We want the computer to predict $y_i$ based on the information of $x_i$, so we want to learn the underlying mapping that maps each $x_i$ to $y_i $. 
To approximate the mapping, we use a neural network $ f_{ \theta }:  \mathbb{R}^{d_x}  \rightarrow  \mathbb{R}^{ d_y }  $, 
which maps an input $x$ to a predicted output $\hat{y}$.
A standard fully-connected neural network is given by
\begin{equation}\label{neural net def}
  f_{\theta} (x) =  W^{ L  } \phi (   W^{ L-1} \dots \phi(  W^2 \phi (W^1 x  )  )  )  ,
\end{equation}
where  
$\phi:  \mathbb{R} \rightarrow  \mathbb{R} $ is the neuron activation function (sometimes simply called ``activation'' or ``neuron''), $W^j $ is a matrix of dimension $d_{j }\times d_{j - 1 } $, $j =  1, \dots, L$ and $\theta = ( W^1, \dots, W^L  )$ represents the collection of all parameters. Here we define $d_0 = d_x$ and $d_L = d_y$. 
When applying the scalar function $\phi$ to a matrix $Z$, we apply $\phi$ to each entry of $Z$. 
Another way to write down the neural network is to use a recursion formula:
\begin{equation}\label{recursive definition of DNN}
	 z^0 = x;    \quad  z^l = \phi( W^l  z^{l -1 } ), \;  l = 1, \dots, L.
\end{equation}
Note that in practice, the recursive expression should be  $ z^l = \phi( W^l  z^{l -1 }  + b^l ) $.
For simplicity of presentation,  throughout the paper, we often skip the ``bias'' term $b^l$ in the expression of neural networks and just use  the simplified version \eqref{recursive definition of DNN}. 

We want to pick the parameter of the neural network so that the predicted output $\hat{y}_i = f_{\theta} (x_i ) $ is close to the true output $y_i$, thus we want to minimize the distance between $y_i$ and $\hat{y}_i$. For a certain distance metric $ \ell(\cdot, \cdot)  $, the problem of finding the optimal parameters can be written as
\begin{equation}\label{main problem}
  	\min_{\theta }  F(\theta) \triangleq  \frac{1}{ n } \sum_{i=1}^n  \ell( y_i,  f_{\theta} ( x_i )  ). 
\end{equation}
For regression problems, $\ell(y, z) $ is often chosen to be the quadratic loss function $\ell(y,z) = \| y - z \|^2$. For binary classification problem, a popular choice of $\ell$ is $\ell (y, z) = \log (1 + \exp ( - y z) ) $. 

Technically,  the neural network given by \eqref{recursive definition of DNN} should be called fully connected feed-forward networks (FCN). 
Neural networks used in practice  often have more complicated structure.
For  computer vision tasks, convolutional neural networks (CNN) are 
standard. 
  In natural language processing, extra layers such as ``attention'' are commonly added. 
Nevertheless, for our purpose of understanding the optimization problem,
we mainly discuss the FCN model \eqref{recursive definition of DNN} throughout this article,
though in few cases the results for CNN will be mentioned.

For a better understanding of the problem \eqref{main problem}, we relate it to several classical optimization problems.

\subsection{Relation with Least Squares}

One special form of \eqref{main problem} is the linear regression problem (least squares):
\begin{equation}\label{formulation: least square problem}
\min_{ w  \in \mathbb{R}^{ d \times 1 } }  \| y - w^T  X  \|^2,    
\end{equation}
where $X = (x_1, \dots, x_n) \in \mathbb{R}^{d \times n} $ , $y \in \mathbb{R}^{1 \times n} $. 
If there is only one linear neuron that maps the input $x$ to $w^T x$ and the loss function is quadratic, then the general neural network problem  \eqref{main problem}  reduces to the least square problem \eqref{formulation: least square problem}.
We explicitly mention the least square problem for two reasons. First, it is one of the simplest forms of a neural network problem.
Second, when understanding neural network optimization, researchers have constantly resorted to insight gained from analyzing linear regression.

\subsection{Relation with Matrix Factorization }

Neural network optimization \eqref{main problem}  is closely related to a fundamental problem in numerical computation: matrix factorization.  If there is only one hidden layer of linear neurons and the loss function is quadratic, and the input data matrix $X $ is the identity matrix,   the problem \eqref{main problem} reduces to
\begin{equation}\label{formulation: MF problem}
\min_{ W_1, W_2 }  \| Y  - W_2 W_1   \|_F^2,    
\end{equation}
where $W_2 \in \mathbb{R}^{d_y \times d_1 } $, $W_1 \in \mathbb{R}^{d_1 \times n } $, $Y  \in \mathbb{R}^{d_y \times n} $ and $\| \cdot   \|_F$ indicates the Frobenious norm of a matrix.
If $d_1 < \min \{  n,  d_y  \}$, then the above problem gives the best rank-$d_1$ approximation of the matrix $Y$. 
Matrix factorization is widely used in engineering, and it has many popular extensions such as non-negative matrix factorization and low-rank matrix completion. Neural network  can be viewed as an extension of two-factor matrix factorization to multi-factor nonlinear matrix factorization.

 \section{ Gradient Descent: Implementation and Basic Analysis}\label{sec: GD}
 
 A large class of methods for neural network optimization are based on gradient descent (GD).
 The basic form of GD is 
\begin{equation}\label{GD general}
   \theta_{t+1}    =    \theta_t - \eta_t  \nabla F(\theta_t),
\end{equation}
 where $\eta_t $ is the step-size (a.k.a. ``learning rate'') and
   $ \nabla F(\theta_t) $ is the gradient of the loss function for the $t$-th iterate. 
 A more practical variant is SGD (Stochastic Gradient Descent):
   at the $t$-th iteration, randomly pick $ i $ and update the parameter by 
 $$
 \theta_{t+1} = \theta_t -  \eta_t \nabla F_i( \theta_t ),
 $$
 where $ F_i(\theta) \triangleq   \ell( y_i,  f_{\theta} ( x_i )  ) $. 
 We will discuss SGD in more detail in Section \ref{sec: general algorithms};
in this section we will only consider simple GD and SGD.

In the rest of the section, we first discuss the computation of the gradient by ``backpropagation'', then discuss classical convergence analysis for GD.  

 \subsection{Computation of Gradient: Backpropagation}

  The discovery of backpropagation (BP) was considered an important landmark in the history of neural networks. From an optimization perspective, it is just an efficient implementation of gradient computation \footnote{While  using GD to solve an optimization problem is straightforward, discovering BP is historically nontrivial. 
 }. 
To illustrate how BP works, suppose the loss function is quadratic and consider the \textit{per-sample} loss of the non-linear network problem
$ F_i (\theta ) =  \| y_i  -   W^L \phi(   W^{L-1} \dots W^2 \phi(  W^1 x_i )  ) \| ^2  .$
The derivation of BP applies to any  $ i $, thus for simplicity of presentation we ignore the
subscript $i$, and use $ x $ and $ y $ instead. 
In addition, to distinguish the per-sample loss with the total loss $F(\theta)$, we use $F_0( \theta  )$
to denote the per-sample loss function:
\begin{equation}\label{back prop nonlinear}
F_0 (\theta ) =  \| y  -   W^L \phi(   W^{L-1} \dots W^2 \phi(  W^1 x )  ) \| ^2 .
\end{equation}

We define an important set of intermediate variables: 
\begin{equation}\label{pre and post-activations}
\begin{split}
 z^0 = x, \quad   &    \quad   h^1 = W^1 z^0,    \\
  \;  z^1 = \phi( h^1 ),  &  \quad    h^2 = W^2 z^1,  \\
    \vdots,   &  \quad   \vdots    \\
    \;  z^{L-1} = \phi( h^{L - 1 } ),  &  \quad    h^L = W^L z^{L-1} .  
\end{split}
\end{equation}
Here,  $h^l $ is often called pre-activation since it is the value that flows into the neuron,
and $z^l$ is called post-activation since it is the value comes out of the neuron. 
Further, define $ D^l = \text{diag}( \phi'(h^l_1), \dots, \phi'(h^l_{d_l }) ) $, which is a diagonal matrix with the $t$-th diagonal entry being the derivative of the activation function evaluated at the $t$-th pre-activation  $ h^l_t$.

Let the error vector $e = 2( h^L - y ) $ \footnote{If the loss function is not quadratic, but a general loss function $ \ell( y,  h^L ) $, we only need to replace $e = 2(h^L - y)$ by  $e = \frac{\partial  \ell  }{ \partial h^L } $.}. 
The gradient over weight matrix $W^l$ is given by 
\begin{equation}\label{nonlinear gradient formula}
\frac{\partial F_0 }{ \partial W^l  } = ( W^L D^{L-1}  \dots W^{l+2} D^{l+1}W^{ l+1 } D^l   )^T  e  (  z^{l-1} )^T, \quad l=1, \dots L.
\end{equation}
Define a sequence of backpropagated error as
\begin{equation}\label{back propogated error}
\begin{split}
e^L        &  = e,      \\
e^{L-1}  &  = (D^{L-1} W^L )^T e^L,  \\
&  \dots,      \\
  e^1       &  = ( D^1 W^2  )^T e^2. 
\end{split}
\end{equation}
Then the partial gradient can be written as
\begin{equation}\label{gradient formula}
\frac{\partial F_0 }{ \partial W^l  } =  e^l  (z^{l-1})^T,   \quad   l = 1, 2, \dots, L.  
\end{equation}

This expression does not specify the details of computation. 
A naive method to compute all partial gradients would require $ O(L^2) $ matrix multiplications
since each partial gradient requires $O(L)$ matrix multiplications.
Many of these multiplication are repeated, thus a smarter algorithm is to reuse the multiplications, similar to the memorization trick in dynamical programming.
More specifically, the algorithm back-propagation  computes all partial gradients in  a forward pass and a backward pass. In the forward pass, from the bottom layer $1$ to the top layer $L$,  post-activation $z^l$ is computed recursively via \eqref{pre and post-activations} and stored for future use.
After computing the last layer output $f_{\theta}(x) =  h^L $, we  compare it with the ground-truth $y$ to obtain the error $e = \ell ( h^L, y)$. 
In the backward pass, from the top layer $L$ to the bottom layer $1$, two quantities are computed at each layer $l$.
First,  the backpropagated error $e^l$ is computed according to \eqref{back propogated error},
i.e., left-multiplying $e^{l+1}$ by the matrix $ ( D^{l-1} W^l )^T   $.
 Second,
 the partial gradient over the $l$-th layer weight matrix $W^l$ is computed by \eqref{gradient formula}, i.e.,
 multiply  the backward signal $e^l$ and  the pre-stored feedforward signal $ ( z^{l-1} )^T $. 
After the forward pass and the backward pass, we have computed the partial gradient for each weight (for  one sample $x$). 

By a small modification to this procedure, we can implement SGD as follows.  
 In the backward pass, for each layer $l$, after computing the partial gradient over $W^l$, we update $W^l$ by a gradient step. 
 After updating all weights $W^l$, we have completed one iteration of SGD. 
In mini-batch SGD, the implementation is slightly different: in the feedforward and backward pass, a mini-batch of multiple samples will pass the network together.

Rigorously speaking, the term ``backpropagation'' refers to algorithm that computes the partial gradients,  i.e., for a mini-batch of samples, computing the partial gradients  in one forward pass and one backward pass.
Nevertheless, it is also often used to describe the entire learning algorithm, especially SGD.

 \subsection{Basic Convergence Analysis of GD}\label{subsec: convergence anlaysis of GD}
 
 In this subsection, we discuss what classical convergence results can be applied to a neural network problem with minimal assumptions. 
 Convergence analysis tailored for neural networks under strong assumptions  will be discussed in Section \ref{sec: global optimization}.
 Consider the following question: 
 \begin{equation}\label{question of convergence}
 \text{    Does gradient descent converge for neural network optimization \eqref{main problem}?  }
 \end{equation}

 \textbf{Meaning of ``convergence''.} 
 There are multiple criteria of convergence. Although we wish
  that the iterates  converge to a global minimum, a more common statement in classical results is  ``every limit point is a stationary point''  (e.g. \cite{bertsekas1997nonlinear}).
  Besides the gap between stationary points and global minima (will be discussed in Section \ref{sec: global optimization}), this claim does not exclude a few undesirable cases: 
  (U1)  the sequence could have more than one limit points; 
 (U2)  limit points could be non-existent \footnote{In logic, the statement ``every element of the set $A$ belongs to the set $B$'' does not imply the set $A$ is non-empty; if the set $A$ is empty, then the statement always holds.
 For example, ``every dragon on the earth is green'' is a correct statement, since no dragon exists. }, i.e.,  the sequence of iterates can diverge.
 Eliminating (U1) and (U2) to ensure convergence to a single stationary point is not easy;
 see Appendix \ref{appen: convergence discussions} for more discussions.

Another criterion is the convergence of function values. 
 This kind of convergence is very easy to achieve:  if the function value is lower bounded by $0$ and the sequence $F(\theta_t)$ is decreasing, then the sequence must converge to some finite value $\hat{F}$. However, optimizers do not regard this as a meaningful criterion since $\hat{F}$ could be an arbitrary value.
 

In this section, we focus on a meaningful and simple convergence criterion: the gradients
of the iterates converge to zero. 
We notice that the objective function $F$ is lower bounded in most machine learning  problems.
Even if (U1) and (U2) happen, classical convergence results do guarantee that  $ \{\nabla F(\theta_t )  \} \rightarrow 0  $ if $F$ is lower bounded.
For many practitioners, this  guarantee is already good enough.

\textbf{Convergence theorems.} 

 There are mainly two types of convergence results for gradient descent. 
 Proposition 1.2.1  in \cite{bertsekas1997nonlinear} applies to the minimization of any differentiable function, but it requires
 line search that is rarely used in large-scale neural network training, so we ignore it. 
 A result more well-known in machine learning area requires  Lipschitz smooth gradient. 
 Proposition 1.2.3 in \cite{bertsekas1997nonlinear}  states that if $ \| \nabla F( w ) - \nabla F(v)  \| \leq \beta \| w - v \| $ for all $w , v$, then for GD with constant stepsize less than $2/\beta $,
 every limit point is a stationary point; further, if the function value is lower bounded,
 then the gradient converges to $0$ \footnote{The convergence of gradient is not stated explicitly in Proposition 1.2.3  of \cite{bertsekas1997nonlinear}, but 
 is straightforward to derive based on the proofs.  }.   
 These theorems require the existence of a global Lipschitz constant $\beta$ of the gradient.
However, for neural network problem \eqref{main problem} a global Lipschitz constant does not exist, thus there is a gap between the theoretical assumptions and the real problems. 
 Is there a simple way to fix this gap?
 
 Unfortunately, for rigorous theoreticians, there seems to be no simple way to fix this gap.
The lack of global Lipschitz constants is a  general challenge for non-linear optimization,
and we refer interested readers to Appendix \ref{appen: convergence discussions} for a 
more in-depth discussion.  
For practitioners, the following claim may be enough 
for a conceptual understanding of the convergence theory: 
if all iterates are bounded, then GD with a proper constant stepsize converges \footnote{This statement is 
	somewhat strange from a theoretical perspective, since we do not know a priori whether the iterates are bounded. However, an assumption of bounded iterates 
	is common in optimization literature. 
 }.  
Bounded Lipschitz constants only help the convergence of the generated sequence, but do not guarantee fast convergence speed.  A more severe issue for the Lipschitz constant is that it may be exponentially large or exponentially small even if it is bounded. 
 In the next section, we will focus on a closely related issue of gradient explosion/vanishing.

\section{ Neural-net Specific Tricks } \label{sec: gradient explosion and vanishing}



Without any prior experience, training a neural network to achieve a reasonable accuracy can be rather challenging. Nowadays, after decades of trial and research, people can train a large network relatively easily (at least for some applications such as image classification). 
In this section, we will describe some main tricks needed for training a neural network.

\subsection{Possible Slow Convergence Due to Explosion/Vanishing}\label{subsec: 1-dim example of grad explosion}

The most well-known difficulty of training deep neural-nets is probably gradient explosion/vanishing.  
A common description of gradient explosion/vanishing is from a signal processing perspective. 
Gradient descent can be viewed as a feedback correction mechanism: the error at the output layer will be propagated back to the previous layers so that the weights are adjusted to reduce the error. Intuitively, when signal propagates through multiple layers, it may get amplified at each layer and thus explode, or get attenuated at each layer and thus vanish. In both cases, the update of the weights will be problematic.

We illustrate the issue of gradient explosion/vanishing via a simple example of 1-dimensional problem:
\begin{equation}\label{1dim problem}
\min_{ w_1, w_2, \dots, w_L  \in \mathbb{R}  }  F(w) \triangleq  0.5 (  w_1 w_2 \dots w_L  - 1  )^2.
\end{equation}

The gradient over $w_i$ is
\begin{equation}\label{partial gradient}
\nabla_{w_i} F =  w_1 \dots w_{i-1} w_{i+1} \dots  w_L  (  w_1 w_2 \dots w_L  - 1  )  = 
w_1 \dots w_{i-1} w_{i+1} \dots  w_L   e   , 
\end{equation}
where $  e = w_1 w_2 \dots, w_L  - 1 $ is the error. 
If all $w_j = 2$, then the gradient has norm $2^{ L -1} | e |$ which is exponentially large; if all $w_j = 1/2$, then the gradient has norm $ 0.5 ^{ L - 1} e $ which is exponentially small.

\begin{figure}
	\centering
	\includegraphics[width=6cm,height=4cm]{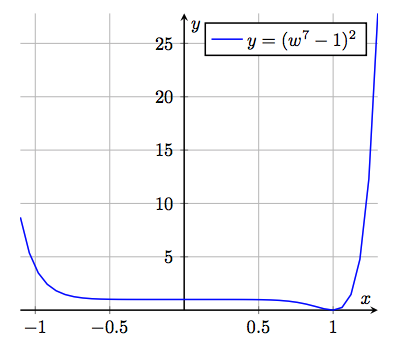}
	\caption{Plot of the function $F(w) = (w^7-1)^2$, which illustrates the gradient explosion/vanishing
	issues. In the region $[-0.8, 0.8]$, the gradients almost vanish;
in the region $[1.2, \infty]$ and $ [ -\infty, -0.8 ] $, the gradients explode.  }
	\label{fig1dPlot}
\end{figure}

\textbf{Example}:  $ F(w) =   (  w^7  - 1  )^2 $, where $w \in \mathbb{R}$ (similar to the example 
analyzed in \cite{shamir2018exponential}).  
This is a simpler version of \eqref{1dim problem}.
The plot of the function is provided in Figure \ref{fig1dPlot}.  The region $[ -1 + c , 1 - c ]$ is flat, which corresponds to vanishing gradient (here $c$ is a small constant, e.g. $0.2$). 
The regions $ [1 + c, \infty] $ and $ [ -\infty, -1 - c ]$ are steep, which correspond to exploding gradient. 
Near the global minimum $ w = 1$, there is a good basin that if initializing in this region GD can converge fast.
If initializing outside this region, say, at $w = -1$, then the algorithm has to traverse the flat region with vanishing gradients which takes a long time.
This is  the main intuition behind  \cite{shamir2018exponential} which proves that it takes exponential time (exponential in the number of layers $L$) for GD with constant stepsize to converge to a global minimum  if initializing near $w_i = -1, \forall i $. 

Theoretically speaking, why is gradient explosion/vanishing a challenge? 
This 1-dimensional example shows that gradient vanishing can make GD with constant stepsize converge very slowly.
In general,  the major drawback of gradient explosion/vanishing  is the slow convergence, 
     due to a large condition number and difficulty in picking a proper step-size. 

  \ifarxiv 
  We remark that gradient explosion and vanishing is often considered to be a more severe issue for recurrent neural network (RNN) than for feed-forward neural networks (see, e.g., Goodfellow et al. \cite{goodfellow2016deep} Sec. 8.2.5), because the same weight matrix is re-used across layers.  
  Another remark is that many works do not mention gradient explosion, but just mention gradient vanishing. This is partially because the non-linear activation function can reduce the signal, and partially because empirical tricks such as ``gradient clipping'' (simply truncating large values in the gradient) can handle gradient explosion to some extent. 
\fi

How to resolve the issue of gradient explosion/vanishing? 
For the 1-dimensional example discussed above,  one can choose an initial point inside the basin near the global minimum. 
Similarly, for a general high-dimensional problem, one solution
is to choose an initial point inside a ``good basin'' that allows the iterates move fast. 

 In the next subsection, we will discuss initialization strategies in detail.

\subsection{Careful  Initialization}
In the rest of this section, we will discuss three major tricks for training deep neural networks. 
In this subsection, we discuss  the first trick: careful initialization. 

As discussed earlier, exploding/vanishing gradient regions indeed exist and occupy a large portion of the whole space, and initializing in these regions will make the algorithm fail. 
 Thus, a natural idea is to pick the initial point in a nice region to start with. 

\textbf{Naive Initialization}
Since the ``nice region'' is unknown, the first thought is to try some simple initial points.
One choice is the all-zero initial point, and another choice is a sparse initial point that
 only a small portion of the weights are non-zero. 
 Yet another choice is to draw the weights from certain random distribution.
 Trying these initial points would be painful as it is not easy to make them always work:
  even if an initialization strategy  works for the current problem,
   it might fail for other neural network problems. 
 Thus, a principled initialization method is needed.

 \textbf{LeCun initialization}
In an early work,  \cite{lecun1998efficient} proposed to initialize a neural network 
  with sigmoid activation functions as follows:
 \begin{equation}\label{Xavier initialization}
E( W^l_{ij}) = 0, \quad   \text{var}( W^l_{ij} ) = \frac{ 1 } { d_{l-1}  }, \quad   l = 1, 2, \dots, L; i = 1, \dots, d_{l-1}; j = 1, \dots, d_l . 
\end{equation}
In other words, the variance of each weight is $1/\text{fan-in}$, where fan-in is the number
of weights fed into the node.  
Although simple, this is a non-trivial finding.
It is not hard to tune the scaling of the random initial point to make it work,
but one may find that one scaling factor  does not work well for another network.
It requires some understanding of neural-nets to realize that adding the dependence on fan-in can lead to a tuning-free initial point. 
 A simple toy experiment can verify the effectiveness of LeCun initialization:
compute the ratio $ \| \Pi_{l = 1}^L W^l x \|/ \| x\|$ for $x = (1;1; \dots ; 1) \in \mathbb{R}^{d \times 1}$ 
  and a random $W \in \mathbb{R}^{d \times d}$ with variance $c$. When $d > 10 L$ and $c = 1/\sqrt{d}$, the ratio is close to $1$;
  when $d > 10 L$ and $ c = 5/\sqrt{d} $ or $ 0.2/\sqrt{d} $, the ratio is very large or small.

 A theoretical  derivation is as follows. 
 Consider a linear neuron with $m$ input $x_1,\dots, x_m$ and one output
  $y = \sum_j w_j x_i  $. Assume the input $x_i $ has zero mean
 and variance $1$, then $y$ has zero mean and variance $ \sqrt{ \sum_{j=1}^m w_j^2  }$.
 To make sure the variance of the output is also $ 1 $, we only need to pick the weights so that
  $ \text{var}(w_j) = 1/m $ and $E[w_j] = 0$. 
The above derivation is for linear activations. 
 If the neuron uses the tanh activation $\phi(t) = \text{tanh}(t) = \frac{2 }{ 1 + e^{- 2 t}} - 1 $, the
gradient $\phi'(t) = \frac{ - 4 e^{-t}  }{ ( 1 + e^{- t} )^2  } $ will be around $ 1 $
in the ``linear regime''.  Thus $ y = \text{tanh}( \sum_j w_j x_j  )$ would have variance approximately
 equal to $1$. 

\textbf{Pre-training and Xavier initialization.}
In late 2000's, the revival of neural networks was attributed to per-training methods that provide good initial point (e.g. \cite{hinton2006reducing, erhan2010does}). 
Partially motivated by this trend, Xavier Glorot and Bengio  \cite{glorot2010understanding} analyzed signal propagation in deep neural networks at initialization, and proposed an initialization method known as Xavier initialization (or Glorot initialization, Glorot normalization): 
\begin{equation}\label{Xavier initialization}
  E( W^l_{ij}) = 0, \quad   \text{var}( W^l_{ij} ) = \frac{ 2 } { d_{l-1} + d_l  }, \quad   l = 1, 2, \dots, L; i = 1, \dots, d_{l-1}; j = 1, \dots, d_l , 
\end{equation}
or sometimes written as   $ \text{var}( W_{ij} ) =  2/ (  \text{fan-in} + \text{fan-out}  ), $
where fan-in and fan-out are the input/output dimensions. 
One example is a Gaussian distribution  $ W^l_{ij}  \sim \mathcal{N}(0,   \frac{ 2 } { d_{l-1} + d_l  } ) $, and another 
example is a uniform distribution 
$ W^l_{ij}  \sim  \text{Unif} [  -   \frac{ \sqrt{6} } { \sqrt{ d_{l-1} + d_l  } }  ,    \frac{ \sqrt{6} } { \sqrt{ d_{l-1} + d_l  } }  ]. $

Xavier initialization can be derived as follows.  For feed-forward signal propagation,
according to the same argument as LeCun initialization,  one could set the  variance
of the weights to be $1/\text{fan-in}$. 
For the backward signal propagation, according to
\eqref{back propogated error},  $ e^{ l } = (  W^{l+1} )^T e^{l+1 }  $ for a linear network.
By a similar argument, one could set the variance of the weights to be $1/\text{fan-out}$.
To handle both feedforward and backward signal propagation, a reasonable heuristic is  to set
$ E( w ) = 0,   \text{var}( w  )  = 2 /  (  \text{fan-in} + \text{fan-out} ) $
for each weight, which is exactly \eqref{Xavier initialization}.

\textbf{Kaiming initialization}. LeCun and Xavier initialization were designed for 
sigmoid activation functions which have slope $1$ in the ```linear regime'' of the activation function. 
 ReLU (rectified linear units) activation \cite{glorot2011deep}  became popular after 2010, and He et al. \cite{he2015delving} noticed that the derivation of Xavier initialization can be modified to better serve ReLU  \footnote{Interestingly,
ReLU was also popularized by Glorot et al. \cite{glorot2011deep}, but they did not apply their own principle to the new neuron ReLU.}.
The intuition is that for a symmetric random variable $ \xi  $, 
 $  E [  \text{ReLU} ( \xi  )  ] =   E [  \max \{    \xi  ,  0  \} ]   = \frac{1}{ 2 }  E[  \xi   ]  $, i.e., 
ReLU cuts half of the signal on average. 
Therefore, they propose a new initialization method
\begin{equation}
	E( W^l_{ij}) = 0, \quad   \text{var}( W^l_{ij} ) = \frac{ 2 } { d_{\text{in}}   }    \text{ or }   \text{var}( W^l_{ij} ) = \frac{ 2 } { d_{\text{out}}   }.
\end{equation}
Note that Kaiming initialization does not try to  balance both feedforward and backward signal propagation like Xavier initialization, but just balances one. 
A recent work \cite{defazio2019scaling} discussed this issue, and proposed and analyzed a geometrical averaging initialization $
\text{var}( w) = c/\sqrt{ ( \text{fan-in} ) \cdot  (  \text{fan-out} )  } 
$
where $c$ is certain constant.

\textbf{LSUV.} Mishkin and Matas \cite{mishkin2015all} proposed layer-sequential unit-variance (LSUV) initialization that consists of two steps: first,  initialize the weights with orthogonal initialization (e.g., see \citet{saxe2013exact}), then for each mini-batch, normalize the variance of the output of each layer to be $1$ by directly scaling the weight matrices. It shows empirical benefits for some problems. 

\textbf{Infinite width networks with general non-linear activations}.  
The derivation of Kaiming initialization cannot be directly extended to general non-linear activations.
Even for one dimensional case where $d_i = 1,  \forall i $, the output of 2-layer neural network $ \hat{y} = \phi(w_2 \phi( w_1 x) )$ for random weights $w_1, w_2 \in \mathbb{R} $ is a complicated random distribution. 
To handle this issue,  Poole et al. \cite{poole2016exponential} proposed to use mean-field approximation to study infinite-width networks. Roughly speaking, based on the central limit theorem that the sum of  a large number of random variables is approximately Gaussian, the pre-activations of each layer are approximately Gaussians, and then they study the evolution of the variance of each layer. 

More specifically, for a given input $x \in \mathbb{R}^{ d_0 } $ and independent random weights $W^1, \dots, W^L$, the pre-activation at each layer $h^l = (h^l_1, \dots, h^l_{d_l})$ are random variables. 
Notice that $ h^l_i = \sum_{ j = 1}^{d_{l-1}}  W^l_{i j } \phi( h^{l-1}_j )  $ is the weighted sum of $d_{l-1}$ 
independent zero-mean random variables $W^l_{i j }$ with weights $\phi( h^{l-1}_j )$. 
As the weights  $\phi( h^{l-1}_j )$ depend on previous layer weights and thus are independent of $W^l_{i j }$, 
one can view the weights as ``fixed''. As the number $d_{l-1}$  goes to infinity, according to central limit theorem,
$ h^l_i $ will converge to a Gaussian distribution with zero mean and a certain variance, denoted as $q^l$. 
For $l \geq 2$,  this variance can be computed recursively as 
\begin{equation}\label{evolution}
q^l = \sigma_w^2  \mathbb{E}_{ \xi \sim \mathcal{N}( 0, 1) }  [ \phi( \sqrt{ q^{l-1} }  \xi  )^2 ] 
=    \sigma_w^2 \int \phi( t \sqrt{q^{l-1}} )^2  \frac{1}{ 2 \pi } \text{exp}( - \frac{t^2}{2} ) dt ,  l = 2, \dots, L , 
\end{equation}
where we assume $E(W_{ij}^l ) = 0$ and $ \text{var}(W^l_{ij}  ) = \frac{ 1 }{d_{ l-1 } } \sigma_w^2   $.
The initial value $q^1$ is  the variance of $h^1_i = \sum_{ j = 1}^{ d_0 }  W^l_{i j } x_j $, which can be computed as  $q^1 = \sigma_w^2 q^0  =\sigma_w^2  \frac{1}{ d_0} \sum_{i=1}^{ d_0 } x_i^2 , $
where $ q^0 = \frac{1}{d_0 } \sum_{i=1}^{d_0 } x_i^2$. 
To check whether $q^l$ computed in such way matches a practical network with finite width, we can compare $q^l$ with the empirical variance $  \hat{q}^l= \frac{1}{ d_l } \sum_{i=1}^{d_l } ( h^l_i )^2  $ to see how close they are.

The evolution equation \eqref{evolution} can be used to guide the design of initialization.
More specifically, for a certain set of $\sigma_w^2$, the equation \eqref{evolution} has a non-zero fixed point $q^*$,
and one can solve the equation numerically and then pick  $q^1 = q^*$ which is achieved by scaling the input vector such that its norm  $\| x \|^2  = q^* d_0 / \sigma_w^2  $. 
Note that for general activation, a case-by-case study of how to pick the initialization variance $\sigma_w^2$
  and the corresponding input norm $ \| x \| $ is needed. 
  See details in  Section 2 of \cite{poole2016exponential}. 
  The practical benefit of such a delicate choice of initial variance  is not clear.
 
  The above discussion is for the bias-free network and only considers the feedforward signal propagation.
  A complete analysis includes the variance of the initial bias as an extra degree of freedom, and the 
  backward signal propagation as an extra equation \footnote{We want to remind the readers
  that Poole et al. \cite{poole2016exponential} derived this extra equation by considering the propagation of the covariance of two inputs (see Section 3 of \cite{poole2016exponential}), while the same equation is presented based on the backward signal propagation in a later paper Pennington et al.
  \cite{pennington2017resurrecting}. Here we recommend viewing the extra equation as backward signal propagation. Interestingly, the NTK paper \cite{jacot2018neural} reviewed later 
  computes the propagation of the covariance of two inputs as well.   
}. 
See more details in  \cite{poole2016exponential} and  \cite{pennington2017resurrecting}. 
\textbf{Finite width networks.}  The analysis of  infinite-width networks can explain the experiments on 
very wide networks, but narrow networks may exhibit different behavior. 
Simple experiments show that the output signal strength will be far from the input signal strength
 if the network is narrow (e.g.   when $d = L $ in the toy experiment described earlier). 

A rigorous quantitative analysis is given in  Hanin and Rolnick \cite{hanin2018start}, which analyzed finite width networks with ReLU activations.
The quantity of interest is $ M_l =  \frac{1}{d_l } \|  z^l \|^2 , l=1, \dots, L$, the normalized post-activation length (this is very similar to $\hat{q}^l $ considered in Poole et al. \cite{poole2016exponential}). 
The previous analysis of \cite{glorot2010understanding} and \cite{he2015delving}
is concerned about the failure mode that the expected output signal strength $\mathbb{E}[M_L ]$ explodes or vanishes. 
\cite{hanin2018start} analyzed another failure mode that the empirical variance across layers
$  \hat{ V } \triangleq  \frac{1}{ L } \sum_{j = 1}^L M_j^2  -  \left(  \frac{1}{ L }  \sum_{j = 1}^{  L } M_j   \right)^2  $
explodes or vanishes. 
They show that with Kaiming initialization (each weight is a zero-mean random variable with variance $2/\text{fan-in}$), the expectation of the empirical variance $ \mathbb{E} [  \hat{ V }  ]   $
is roughly in the order of $ \text{exp}\left( \sum_{k=1}^L   \frac{1}{ d_k } \right).  $
If all layers have the same width $d$, then $ \mathbb{E} [  \hat{ V }  ]   $ is in
the order of $ \text{exp}(L/d ) $. Therefore, for fixed width $d$, increasing the depth $L$ can make
the signal propagation unstable. 
This might be helpful for explaining why training deep networks is difficult (note that
there are other conjectures on the training difficulty of deep networks; e.g. \cite{orhan2017skip}). 

\textbf{Dynamical isometry.}
Another line of research that aims to understand signal propagation is based on the notion
of dynamical isometry  \cite{saxe2013exact}.  
It means that the input-output Jacobian (defined below)  has \textit{all} singular values close to $1$.  Consider a neural-net $ f(x) = \phi( W^L \phi(  W^{L-1 } \dots  \phi(W^1 x ) ) ) $, which is slightly
different from \eqref{neural net def} (with an extra $\phi$ at the last layer). Its ``input-output Jacobian''
 is
 $$ \frac{ \partial z^L }{\partial z^0}  = \Pi_{l = 1}^L ( D^l W^l) , $$
 where $D^l$ is a diagonal matrix with entries being the elements of $\phi'( h^l_1, \dots, h^l_{d_l } )$.
 If all singular values of $J$ are close to $1$, then according to 
 \eqref{back propogated error}, the back-propagated error $e^l, l=1, \dots, L$ will
 be of similar strength. 

Achieving  isometry in deep  linear networks with equal width (i.e. all $d_l$'s are the same) is very simple: 
just picking each $W^l$ to be an orthogonal matrix, then their product $W^L W^{L-1} \dots W^1 $
is  an orthogonal matrix and thus has all singular values being exactly $1$. 
Saxe et al. \cite{saxe2013exact} showed empirically that for deep linear networks, this orthogonal initialization leads to depth-independent training time, while Gaussian initialization cannot achieve depth-independent training time.
This seems to indicate that orthogonal initialization is better than Gaussian initialization,
but for non-linear networks this benefit was not observed. 

Later, a formal analysis for deep non-linear networks with infinite width was provided in Pennington et al. \cite{pennington2017resurrecting, pennington2018emergence}. They  used tools from free probability theory to compute the distribution of all singular values of the input-output Jacobian 
(more precisely, the limiting distribution as the width goes to infinity).
An interesting discovery is that dynamical isometry can be achieved  when using  sigmoid activation and orthogonal initialization, but cannot be achieved for Gaussian initialization. 
Note that one needs to carefully pick  $\sigma_w^2$, $\sigma_b^2$ and $\| x\|^2$,
and simply using orthogonal initialization is not enough, which partially explains
why Saxe et al. \cite{saxe2013exact}  did not observe the benefit of orthogonal initialization.

\textbf{Dynamical isometry for other networks.}
One  obstacle of applying orthogonal initialization to practical networks is 
convolution operators: it is not clear at all how to compute an ``orthogonal'' convolution operator. 
Xiao et al. \cite{xiao2018dynamical} further studied how to achieve dynamical isometry in deep CNN.
They proposed two orthogonal initialization methods for CNN (the simpler version is called DeltaOrthogonal), with which they can train a 10000-layer CNN without other tricks like  batch-normalization or skip connections (these tricks are discussed later). This  indicates that for training ultra-deep networks, carefully chosen initialization is enough (note that the test accuracy on CIFAR10 is not as good as state-of-the-art perhaps due to 
the limited representation power around that initial point). 

The analysis of dynamical isometry has been applied to other neural networks as well. 
\citet{li2018on} analyzed dynamical isometry for deep autoencoders, 
and showed that it is possible to train a 200-layer autoencoder without tricks like layer-wise pre-training and batch normalization.
\citet{gilboa2019dynamical} analyzed dynamical isometry for LSTM and RNNs, 
and proposed a new  initialization scheme that performs much better than traditional initialization schemes in terms of reducing training instabilities. 

\textbf{Computing spectrum.} Empirically computing the spectrum (of certain matrices)
is very useful for understanding the training process. 
Dynamical isometry is about the input-output Jacobian, and there are a few other
matrices that have been studied. 
\citet{sagun2016singularity, sagun2017empirical} plotted the distribution of eigenvalues of the Hessian for shallow neural networks. They observed a few outlier eigenvalues that are a few orders of magnitudes larger than other eigenvalues. 
Computing the eigenvalues for large networks is very time consuming.  To tackle this challenge, \citet{ghorbani2019investigation} used stochastic Lanczos quadrature algorithm
to estimate the spectrum density for large-scale problems such as 32-layer ResNet on ImageNet.
They confirmed the finding of outlier eigenvalues of \cite{sagun2017empirical} for
 ImageNet. 
Besides the details of the eigenvalue distributions, these numerical findings
indeed verify that the local Lipschitz constant of the gradient (the maximum eigenvalue of the Hessian)   is rather small in practical neural network training, partially  due to careful initialization. 


Along a different line,   \citet{brock2018large} calculated the top three eigenvalues of the weight matrices (not the Hessian) to track the training process of generative adversarial networks. 
In a convolutional neural network,  the weight is actually a tensor, and \citet{brock2018large} reshaped it into a matrix and computes the spectrum of this matrix. 
  \citet{sedghi2018the} provided a simple formula to exactly compute the singular values of
the linear transformation of each layer, which is defined by a convolution operator.

\subsection{Normalization Methods}

The second approach is  normalization during the algorithm.
This can be viewed as an extension of the first approach: instead of merely modifying the initial point, this approach modifies the network for all the following iterates.
One representative method is  batch normalization (BatchNorm) \cite{ioffe2015batch}, which is a standard technique nowadays. 

\textbf{Preparation: data normalization.}
To understand BatchNorm, let us first review a common data preprocessing trick: for linear regression problem $\min_w  \sum_{i = 1}^n (y_i  - w^T  x_i  )^2 $,
we often scale each row of the data matrix $[x_1, x_2, \dots, x_n ] \in \mathbb{R}^{d_x \times n} $
so that each row has zero mean and unit norm (one row corresponds to one feature).
This operation can be viewed as a pre-conditioning technique
that can reduce the condition number of the Hessian matrix,
which increases the convergence speed of gradient-based methods. 

\textbf{Motivation of BatchNorm: layerwise normalization.}
How to extend this idea to deep neural-nets?
Intuitively, the convergence speed of each weight matrix $W^l$ is related to the ``input matrix'' to that layer, which is the matrix of pre-activations  $[h^l (1), h^l(2), \dots, h^l(n) ]$, where $h^l(k)$ represents the pre-activation at the $l$-th layer
 for the $k$-th sample ($h^l$ is defined in \ref{pre and post-activations}). 
 Thus it is natural to hope that each row of $[h^l (1), h^l(2), \dots, h^l(n) ]$
 has zero mean and unit variance. 
 To achieve the extra goal,  a naive method is to normalize the pre-activation matrix after updating all weights by a gradient step, but it may ruin the convergence of GD.
 
 \textbf{Essence of BatchNorm.}
 The solution of \cite{ioffe2015batch} is to view this normalization step as a nonlinear transformation ``$\text{BN}$'' and add BN layers to the original neural network. BN layers play  the same role
 as  the activation function $\phi  $ and other layers (such as max pooling layers).
 This modification can be consistent with BP as long as the chain rule of the gradient can be applied, or equivalently,  the gradient of this operation BN can be computed.
Note that a typical optimization-style solution would be to add constraints that encode
the requirements; in contrast, the solution of BN is to add a non-linear transformation
to encode the requirements.  This is a typical neural-net style solution. 
More details of BatchNorm are given in  Appendix  \ref{appen: batch norm}. 

\textbf{Understanding BatchNorm}.
The original BatchNorm paper claims that BatchNorm reduces the ``internal covariate shift''.
\citet{santurkar2018does} argues that  internal covariate shift has little do with the success
of BatchNorm, and the major benefit of BatchNorm is to reduce the Lipschitz constants (of
the objective and the gradients). 
\citet{bjorck2018understanding}  shows that the benefit of BatchNorm is to allow larger 
learning rate, and discusses the relation with initialization schemes. 
\citet{arora2018theoretical, cai2019quantitative, kohler2019exponential} 
analyzed the theoretical benefits of BatchNorm (mainly larger or auto-tuning learning rate) under various settings. 
\citet{ghorbani2019investigation}  
numerically found that for networks without BatchNorm, there are large isolated eigenvalues, while for networks with BatchNorm this phenomenon does not occur.

\textbf{Other normalization methods}.
One issue of BatchNorm is that the mean and the variance for each mini-batch is
computed as an approximation of the mean/variance for all samples,
thus if different mini-batches do not have similar statistics then BN does not work very well. 
Researchers have proposed other normalization methods such as  weight normalization \cite{salimans2016weight},  layer normalization \cite{ba2016layer}, instance normalization 
\cite{ulyanov2016instance}, group normalization \cite{wu2018group} and spectral normalization \cite{miyato2018spectral} and switchable normalization \cite{luo2019switchable}.

These methods can be divided into two classes.
The first class of methods  normalize the intermediate outcome of the neural network (often the pre-activations). For a pre-activation matrix $ (h(1), \dots, h(n)) $ at a certain layer (we ignore the layer index), BatchNorm chooses to normalize the rows (more precisely, divide each row into many segments and normalize each segment), layer normalization  normalizes the columns, and group normalization  normalizes a sub-matrix that consists of a few columns and a few rows. 

The second class of methods directly normalize the weight matrices. 
Weight normalization \cite{salimans2016weight} reparameterizes a weight vector $w$
as $ g \frac{v }{ \| v \| } $, i.e. separates the norm and the direction of the weight matrix, and solve a
new problem with $g$ and $v$ being new parameters to learn. 
Spectral normalization \cite{miyato2018spectral} changes the weight matrix $W$ to
$\text{SN}(W) = \frac{ W }{ \sigma_{\max}(W) }  $ where $\sigma_{\max}(W) $ is the spectral norm
of $ W $, and considers a new neural network
$ f_{\theta }(x) =  \text{SN}(W^L) \phi ( \text{SN}(W^{L-1}) \dots \text{SN}(W^{2})  \phi ( 
\text{SN}(W^{1}) x ) \dots  ) .$
Some of these normalization methods can outperform BatchNorm in a few scenarios, such as RNNs \cite{ba2016layer}, problems where only small mini-batches are available  \cite{wu2018group} and generative adversarial networks \cite{miyato2018spectral}.


 \subsection{Changing Neural Architecture}
The third approach  is to change the neural architecture. 
Around 2014, people noticed that from AlexNet \cite{krizhevsky2012imagenet} to Inception \cite{szegedy2015going}, the neural networks get deeper and the performance gets better, thus it is natural to further increase the depth of the network.
However, even with smart initialization and BatchNorm, people found training more than 20-30 layers
is very difficult. 
As shown in \cite{he2016deep},  for a given network architecture VGG,  
a 56-layer network achieves worse training and test accuracy than a 20-layer network \footnote{Note that this difficulty is probably not due to gradient explosion/vanishing, and perhaps related
to singularities \cite{orhan2017skip}.  
}.
Thus, a major challenge at that time was to make training an ``ultra-deep'' neural network possible.

\textbf{ResNet.}
The key trick of ResNet \cite{he2016deep} is simple: adding an identity skip-connection for every few layers. 
 More specifically, ResNet changes the network from \eqref{recursive definition of DNN} to 
\begin{equation}\label{recursive definition of ResNet}
 z^0 = x;  z^l =  \phi(   \mathcal{F} (W^l, z^{l-1} ) + z^{l-1} ),  \;  \quad   l = 1, \dots, L, 
 \end{equation} 
 where $\mathcal{F}$ represents a few layers of the original networks, such as $ \mathcal{F} (W_1, W_2, z )   = W_1 \phi( W_2 z ) .  $ 
 Note that a commonly seen expression of ResNet (especially in theoretical papers) is 
 $  z^l =    \mathcal{F} (W^l, z^{l-1} ) + z^{l-1}   $, which does not have the extra $\phi(\cdot)$,
 but \eqref{recursive definition of ResNet} is the form used in practical networks. 
 Note that the expression \eqref{recursive definition of ResNet} only holds 
 when the input and output have the same dimension; to change the dimension across layers,
 one could use extra projection matrices (i.e. change the second term $z^{l-1}$ to $ U^l z^{l-1} $)
 or use other operations (e.g. pooling). 
In theoretical analysis,  the form of \eqref{recursive definition of ResNet} is often used. 
 
 
 ResNet has achieved remarkable success:  
 with the simple trick of adding identity skip connection (and also BatchNorm),
  ResNet with 152 layers greatly improved the best test accuracy at that time 
  for a few computer vision tasks including  ImageNet classification (improving top-5 error to a remarkable 3.57$\%$).

\textbf{Other architectures.}
Neural architecture design is one of the major threads of current deep learning research.
Other popular architecture related to ResNet include
high-way networks \cite{srivastava2015highway},  DenseNet
\cite{huang2017densely} and ResNext \cite{xie2017aggregated}.
While these architectures are designed by humans, another recent trend is 
the automatic search of neural architectures (neural architecture search) \cite{zoph2016neural}. 
There are also intermediate approaches: search one or few hyper-parameters of the neural-architecture
such as the width of each layer \cite{yu2019network,tan2019efficientnet}.
Currently, the state-of-the-art architectures (e.g. EfficientNet \cite{tan2019efficientnet}) for ImageNet classification can achieve much higher top-1 accuracy than ResNet (around 85$\%$
v.s. 78 $\%$) with the aid of a few extra tricks. 

\textbf{Analysis of ResNet and initialization.}
Understanding the theoretical advantage of ResNet or skip connections has attracted much attention. 
 The benefits of skip connections are likely due to multiple factors, including better generalization ability (or feature learning ability), better signal propagation and better optimization landscape.
For instance, \citet{orhan2017skip} suggests that skip connections improve the landscape
by breaking symmetry. 

Following the theme of this section on signal propagation, we discuss some results on the signal propagation aspects of ResNet.
As mentioned earlier, \citet{hanin2018neural} discussed two failure modes for training; in addition, it proved  that for ResNet if failure mode 1 does not happen then failure mode 2 does not happen either.  
\citet{tarnowski2019dynamical} 
 proved that for ResNet,  dynamic isometry can be achieved for any activation (including ReLU) and any bi-unitary random initialization (including Gaussian and Orthogonal initialization). In contrast, for the original (non-residual) network, dynamic isometry is achieved only for orthogonal initialization and certain activations (excluding ReLU). 

Besides theoretical analysis, some works further explored the design of new initialization schemes
such as  \cite{yang2017mean,balduzzi2017shattered,zhang2019fixup}. 
\citet{yang2017mean} analyzed randomly initialized ResNet and 
  showed that the optimal initial variance is different from Xavier or He initialization and should
 depend on the depth.
 \citet{balduzzi2017shattered} analyzed  ResNet with recursion $ z^{l+1} =  z^l  + \beta  W^l  \cdot  \text{ReLU}(z^{l}) $, where $\beta$ is a scaling factor. 
It showed that for $\beta$-scaled ResNet with BatchNorm and Kaiming intialization,  the correlation of two input vectors scales as   $ \frac{1}{ \beta \sqrt{L}} $, thus  it suggests a scaling factor $\beta  = 1/\sqrt{L} $. 
\citet{zhang2019fixup} analyzed the signal propagation of ResNet carefully, and proposed Fixup initialization which leads to good performance on ImageNet, without using BatchNorm.  
This is probably the first such good result on ImageNet without normalization methods. 
It modifies Kaiming initialization in the following ways:
first, scale all weight layers inside residual branches by $ L^{ - 1/(2 m - 2) }$,
where $ m $ is the depth of each residual branch and $L$ is the number of ``residual layers''
 (e.g. for ResNet50,  $m = 3$, $L = 16$);  second, set the last layer of each residual branch to $0$;
 third, add a scalar multiplier and bias to various layers. 
 
 The major modification of Fixup initialization  is the scaling factor $ L^{ - 1/(2 m - 2) }$, and the intuition
 can be understood by the following simple examples. 
Consider a linear 1-dimensional ResNet $ y = ( 1 + w_L  ) \dots (1 + w_2) (1 + w_1 ) x $  where the scalars  $w_i \sim \mathcal{N}(0, c  )$.  To  ensure that $ \| y \| /  \|  x \| \approx O(1) $, 
we need to pick $c \leq 1/L $. Note that if $c \approx  0$, of course $\| y \|/ \|x  \| \approx  1 $, 
but then the network has little representation power, thus we want to pick  $c$ as large as possible, such as $c = 1/L$. This explains the  part of $L^{-1}$ in the scaling factor.
Consider another 1-dimensional ResNet  $ y  =  (1 +  u_m^L \dots u_2^L  u_1^L ) \dots (1 + u_m^1  \dots u_2^1 u_1^1 ) x  $, where each residual branch has $m$ layers. 
We want $ \text{var}( u_m^i  \dots u_2^i  u_1^i )  = 1/L  $, 
thus it is natural to choose $ \text{var}(u^i_j ) = L^{ - 1/m } $,
or similarly, multiplying a standard Gaussian variable by $ L^{ -1/{2 m } } . $
This is very close to the scaling factor $ L^{ - 1/(2 m - 2) }$ used in Fixup initialization.

 








\subsection{Training Ultra-Deep Neural-nets}

There are a few approaches that can currently train very deep networks (say, more than 1000 layers) nowadays to reasonable test accuracy for image classification tasks. 
\begin{itemize}
	\item  The most well-known approach uses all three tricks discussed above (or variants):  
	proper initialization, proper architecture (e.g. ResNet) and BatchNorm. 
	
	\item  As mentioned earlier, only using a very carefully chosen initial point \cite{xiao2018dynamical}  is enough for training 
	ultra-deep CNNs (though this work does not achieve the best test accuracy). 
	
	\item Using FixUp initialization and ResNet  \footnote{Note that this paper also uses a certain scalar normalization trick that is much simpler than BatchNorm.} \cite{zhang2019fixup}.
\end{itemize}


Besides the three tricks discussed in this section, there are
 quite a few design choices that are probably important for achieving good performance of neural networks. 
These include but not limited to data processing (data augmentation, adversarial training, etc.),  optimization methods (optimization algorithms, learning rate schedule, learning rate decay, etc.),
 regularization ($\ell_2$-norm regularization, dropout, etc.),
 neural architecture (depth, width, connection patterns, filter numbers, etc.)
 and activation functions (ReLU, leaky ReLU, ELU, tanh, swish, etc.). We have only discussed three major design choices which are relatively well understood in this section. 
 We will discuss a few other choices in the following sections, mainly the optimization
 methods and the width. 
 

 
 




\section{General Algorithms for Training  Neural Networks}\label{sec: general algorithms}

In the previous section, we discussed  neural-net specific  tricks.
These tricks need to be combined with an optimization algorithm such as SGD,
and are  largely orthogonal to optimization algorithms. 
 In this section, we discuss optimization algorithms used to solve neural network problems,
 which are often generic  and can be applied to other  optimization problems as well. 

The goals of  algorithm design for neural-net optimization are at least two-fold: 
first,  converge faster; second,  improve certain metric of  interest. 
The metrics of interest can be very different from the optimization loss, and
 is often measured on unseen data.
A faster method does not necessarily generalize better,
and not necessarily improves the metric of interest. 
Due to this gap, a common algorithm design strategy is: try an optimization idea to improve the convergence speed, but only accept the idea if it passes a certain "performance check''. 
In this section, 
we discuss optimization algorithms commonly used in deep learning, 
which are popular due to both optimization reasons  and non-optimization reasons.  
For a more detailed tutorial of standard methods for machine learning (not just deep learning), see Bottou, Curtis and Nocedal \cite{bottou2018optimization} and  \citet{curtis2017optimization} 

 \subsection{ SGD and learning-rate schedules}

We can write \eqref{main problem} as a finite-sum optimization problem: 
\begin{equation}\label{main problem}
\min_{\theta }  F(\theta) \triangleq \frac{1}{ B }  \sum_{i=1}^B  F_i(\theta ).
\end{equation}
Each $F_i(\theta)$ represents the sum of training loss for a mini-batch of training samples (e.g. $32$, $64$ or $512$ samples), and $ B $ is the total number of mini-batches (smaller than the total number of training samples $n$). The exact expression of $F_i$ does not matter in this section, as we only need to know how to compute the gradient  $\nabla F_i(\theta)$.  

Currently, the most popular class of methods are SGD and its variants. 
Theoretically, SGD works as follows: at the $t$-th iteration, randomly pick  $ i $ and update the parameter by 
$$
\theta_{t+1} = \theta_t - \alpha_t  \nabla F_i( \theta_t ). 
$$
In practice, the set of all samples are randomly shuffled at the beginning of each epoch,
then split into multiple mini-batches. At each iteration, one mini-batch is loaded into the memory  for computation  (computing mini-batch gradient and performing weight update).

\textbf{Reasons for SGD: memory constraint and faster convergence}.
The  reasons of using SGD instead of GD are the memory constraint and the faster convergence.
A single GPU or CPU cannot load all samples into its memory for computing the full gradient,
thus loading a mini-batch of samples at each iteration is a reasonable choice. 
Nevertheless, even with the memory constraint, the original GD can be implemented
by accumulating all mini-batch gradients without updating the parameters at each iteration. 
Compared to GD implemented in this way,
the advantage of SGD  is the faster convergence speed. 
We defer a more rigorous description to Section \ref{subsec: theory of sgd}.
We emphasize that SGD is not necessarily faster than GD if all samples can be processed in a single machine in a parallel way, but in the memory-constraint system SGD is often much faster than GD. 

 How strict is the  memory constraint in practice? 
The number of samples in one mini-batch depends on the size of the memory,
 and also depends on the number of parameters in the model and other algorithmic requirement (e.g.
 intermediate output at each layer).  For instance, a GPU with memory size 11 Gigabytes  can only process 512 samples at one time  when using AlexNet for ImageNet, and can only process 64 samples at one time when using ResNet50 for ImageNet \footnote{Most
 implementations of ResNet50 only  process 32 samples in one GPU.}.
 Note that the memory constraint only implies that ``processing mini-batches separately''
 is crucial, but does not imply using gradient methods is crucial.
 The comparison of SGD over other stochastic methods (e.g. stochastic second-order methods)
  is still under research; see Section \ref{subsec: Other algorithms}.

\textbf{Vanilla learning rate schedules}.
Similar to the case in general nonlinear programming, the choice of step-size (learning rate) is 
also important in deep learning. 
In the simplest version of SGD,  constant step-size  $\alpha_t = \alpha $  works reasonably well: it can achieve a very small training error and 
relatively small test error for many common datasets. 
A more popular version of SGD is to divide the step-size  by a fixed constant once every few epochs (e.g. divide by 10 every 5-10 epochs)
or divide by a constant when stuck. 
Some researchers refer to SGD with such simple steps-size update rule as "vanilla SGD".

\textbf{Learning rate warmup}.
``Warmup'' is a commonly used heuristic in deep learning. It means to use a very small
learning rate for a number of iterations, and then increases to the ``regular'' learning rate.
It has been used in a few major problems, including  ResNet \cite{he2016deep}, 
 large-batch training for image classification  \cite{goyal2017accurate},
 and many popular natural language architectures such as Transformer networks
\cite{vaswani2017attention} BERT \cite{devlin2018bert}. 
See \citet{gotmare2018a} for an empirical study of warmup. 

\textbf{Cyclical learning rate}.
A particularly useful variant is SGD with cyclical learning rate (\cite{smith2017cyclical, loshchilov2016sgdr}).
The basic idea is to let the step-size bounce between a lower threshold and an upper threshold.  
In one variant called SGDR (Smith \cite{smith2017cyclical}), 
the general principle is to  gradually  decrease and then gradually increase step-size within one epoch, and one special  rule is to use piecewise linear step-size. A later work \cite{smith2017super} reported  ``super convergence behavior'' that SGDR converges several times faster than SGD in image classification. 
In another variant of Ioshchilov et al. \cite{loshchilov2016sgdr}, within one epoch the step-size gradually decreases to the lower threshold and \textit{suddenly} increases to the upper threshold ("restart"). This ``restart'' strategy resembles classical optimization
tricks in, e.g., Powell \cite{powell1977restart} and O’Donoghue and Candes \cite{o2015adaptive}. 
\citet{gotmare2018a} studied the reasons of the success of cyclical learning rates, but a 
thorough understanding remains elusive. 

\subsection{Theoretical analysis  of SGD}\label{subsec: theory of sgd}

In the previous subsection, we  discussed the  learning rate schedules used in practice;
next, we discuss the theoretical analysis of SGD. The theoretical convergence of SGD has been studied for decades (e.g., \cite{luo1991convergence}).
For a detailed description of the convergence analysis of SGD, we refer the readers 
to \citet{bottou2018optimization}. 
However, there are at least two  issues of the classical analysis. 
First, the existing analysis assumes Lipschitz continuous gradients similar
to the analysis of GD, which cannot be easily justified as discussed in Section \ref{subsec: convergence anlaysis of GD}.
We put this issue aside, and focus on the second issue that is specific to SGD. 

\textbf{Constant v.s. diminishing learning rate} The existing convergence analysis of SGD often requires  diminishing step-size , such as $\eta_t = 1/t^{\alpha}$ for $\alpha \in ( 1/2, 1 ] $
  \cite{luo1991convergence, bottou2018optimization}.
Results for SGD with constant step-size also exist (e.g., \cite[Theorem 4.8]{bottou2018optimization}), but  the gradient does not converge to zero since there is an extra error term dependent on the step-size.
This is because SGD with constant stepsize may  finally enter a ``confusion zone'' in which iterates jump around  \cite{luo1991convergence}.
Early works in deep learning (e.g. \citet{lecun1998efficient}) suggested
 using diminishing learning rate such as $ O (1/t^{0.7})$, but nowadays
   constant learning rate works quite well in many cases. 
For practitioners, this unrealistic assumption on the learning rate makes it harder
 to use the theory to guide the  design of the optimization algorithms.
For theoreticians, using diminishing step-size may lead to a convergence rate far from practical performance.

\textbf{New analysis for constant learning rate: realizable case.}
Recently, an explanation of the constant learning rate has become increasingly popular:  if the problem is realizable (the global optimal value is zero),  then SGD with constant step-size does converge \cite{schmidt2013fast,vaswani2018fast} \footnote{
	Rigorously speaking, the  conditions are stronger than realizability (e.g. weak growth condition  in \cite{vaswani2018fast}). For certain problems such as least squares, realizablity is enough
since it implies the weak growth condition in \cite{vaswani2018fast}. }.
In other words, if the network is powerful enough to represent the underlying function, then the stochastic noise causes little harm in the final stages of training, i.e., realizability has an ``automatic variance reduction'' effect \cite{liu2018mass}. 
Note that ``zero global minimal value'' is a strong assumptions for a general unconstrained
optimization problem, but  the purpose of using neural networks
is exactly to have strong representation power, 
thus ``zero global minimal value''  is a reasonable assumption in deep learning.
This line of research indicates that neural network optimization has special structure, 
thus classical optimization theory  may not provide the best explanations for neural-nets. 

\textbf{Acceleration over GD.}
We  illustrate why SGD is faster than GD by a simple realizable problem. 
Consider a least squares problem $ \min_{w \in \mathbb{R}^d }  \frac{1}{2 n}
\sum_{i=1}^n  (y_i - w^T x_i )^2   $,
and assume the problem is realizable, i.e., the global minimal value is zero. 
For simplicity, we assume $n \geq d$, and  the data are normalized such that $\| x_i \| = 1, \forall i$. 
It can be shown (e.g.  \cite[Theorem 4]{vaswani2018fast}) that
the convergence rate of  SGD with learning rate $\eta = 1$ is $
\frac{n }{ d }  \frac{  \lambda_{\max}  }{     \lambda_{\rm avg}} $ times better than GD, where $\lambda_{\max}$ is the maximum eigenvalue of the Hessian matrix $\frac{1}{n} XX^T $ and $\lambda_{\rm avg}$ is the average eigenvalue of the same matrix. Since $ 1 \leq \frac{  \lambda_{\max}  }{     \lambda_{\rm avg}}  \leq  d $, the result implies that SGD is 
$n/d $ to $ n$ times faster than GD. 
In the extreme case that all samples are almost the same, i.e., $x_i \approx x_1, \forall \; i$,  SGD is about $ n$ times faster than GD.   
In the above analysis, we assume each mini-batch consists of a single sample.
When there are $N $ mini-batches, SGD is roughly $1$ to $N$ times faster than GD. 
In practice,  the acceleration ratio of SGD over GD depends on many factors, and the above analysis can only provide some preliminary insight  for understanding the advantage of SGD.

\subsection{Momentum and accelerated SGD}\label{subsec: Adaptive Methods}  %
Another popular class of methods are SGD with momentum and SGD with Nesterov momentum. SGD with momentum works as follows: at the $t$-th iteration, randomly pick $i$ and update the momentum term and the parameter by
$$
   m_t =  \beta  m_{t-1}  +  (1 - \beta )  \nabla F_i( \theta_t );  \quad 
   \theta_{t+1} = \theta_t -  \alpha_t m_t.  
$$
We ignore the expression of SGD with Nesterov momentum (see, e.g., \cite{ruder2016overview}). 

They are the stochastic versions of the heavy-ball method and accelerated gradient method, but are commonly rebranded as ``momentum methods'' in deep learning. 
 They are widely used in machine learning area 
not only because of  faster speed than vanilla SGD in practice, but also because of the theoretical advantage for convex  or quadratic problems; see Appendix \ref{appen: convergence discussions}
for more detailed discussions. 

\textbf{Theoretical advantage of SGD with momentum.}
The classical results on the benefit of momentum 
 only apply to the batch methods (i.e. all samples are used at each iteration).
 It is interesting to understand whether momentum can improve the speed of the stochastic
 version of GD in theory. 
Unfortunately, even for convex problems, achieving such a desired acceleration is not easy according to various negative results (e.g. \cite{devolder2014first,devolder2013first, kidambi2018insufficiency}). 
For instance, Kidambi et al. \cite{kidambi2018insufficiency}   showed that there are simple quadratic problem instances that momentum does not improve the convergence speed of SGD.
Note that this negative result of \cite{kidambi2018insufficiency} only applies to the naive combination of SGD and momentum terms for a general convex problem.

There are two ways to obtain better convergence rate than SGD. 
First,  by exploiting tricks such as variance reduction,
more advanced optimization methods  (e.g. \cite{lin2015universal, allen2017katyusha}) can achieve an improved  convergence rate
that combines the theoretical improvement  of both momentum and SGD.
However, these methods are somewhat complicated, and are not that popular in practice. 
 \citet{defazio2019ineffectiveness} analyzed the reasons why variance reduction
 is not very successful in deep learning. 
 Second, by considering more structure of the problem, simpler variants of SGD
 can achieve acceleration. 
 \citet{jain2017accelerating} incorporated statistical assumption of the data to 
  show that a certain variant is faster than SGD. 
\citet{liu2018accelerating} considered  realizable quadratic problems,
 and proposed a modified version of SGD with Nesterov's momentum  which is faster than SGD. 

\textbf{Accelerated SGD for non-convex problems.}
The above works only apply  to convex problems and are thus not directly  applicable to neural network problems which are \textit{non-convex}. Designing accelerated algorithms for general non-convex problems is quite hard:  even for the batch version, 
accelerated gradient methods  cannot achieve better convergence rate
than GD when solving non-convex problems.
 There have been many recent works that design new methods with faster convergence rate than
SGD on general non-convex problems (e.g. \cite{carmon2018accelerated, carmon2017convex, xu2018first, fang2018spider, allen2018natasha} and references therein). These methods are mainly theoretical and not yet used by practitioners in deep learning area.
One possible reason is that  they are designed for worst-case non-convex problems, and do not capture the structure of neural network optimization. 




\subsection{Adaptive gradient methods: AdaGrad, RMSProp, Adam and more}

The third class of popular methods are adaptive gradient methods, such as AdaGrad \cite{duchi2011adaptive}, RMSProp \cite{tieleman2012lecture} and Adam \cite{kingma2014adam}. 
We will present these methods and discuss their empirical performance and the theoretical results. 

 
 \textbf{Descriptions of  adaptive gradient methods.}
AdaGrad works as follows: at the $t$-th iteration, randomly pick $i $, and update the parameter as (let $\circ$ denote entry-wise product)
\begin{equation}\label{AdaGrad}
  \theta_{t+1} = \theta_t -  \alpha_t   v_t^{-1/2} \circ g_t , \quad  t= 0, 1, 2, \dots, 
\end{equation}
where $g_t = \nabla F_i ( \theta_t ) $ and  $v_t = \sum_{j = 1}^t  g_j \circ g_j   $. In other words, the step-size for the $k$-th coordinate  is adjusted from $ \alpha_t $ in standard SGD to $ \alpha_t / \sqrt{  \sum_{j = 0}^t  g_{j,k}^2   }  $ where $g_{j,k} $ denotes the $k$-th entry of $g_j$. 
AdaGrad can be also written in the form of a stochastic diagonally scaled GD as
$$
\theta_{t+1} = \theta_t -  \alpha_t   D_t^{-1}  g_t ,
$$
where $D_t = \text{diag} \left(   \sum_{j = 1}^t  g_j  g_j^T   \right)   $ is diagonal part of the matrix formed by the average of the outer product of all past gradients. 
This can be viewed as a stochastic version of the general gradient method in \cite[Section 1.2.1]{bertsekas1997nonlinear}, with a special choice of the diagonal scaling matrix $D_t$.  
AdaGrad is shown to exhibit a convergence rate similar to SGD for convex problems \cite{duchi2011adaptive} and non-convex problems (see, e.g., \cite{chen2018convergence}): when the stepsize is chosen to be the standard diminishing stepsize (e.g. $1/\sqrt{t} $) the iteration complexity is $O( \log T / \sqrt{T})$ (i.e. after $T$ iterations, the error is of the order $1/\sqrt{T} $). 

One drawback of AdaGrad is that it treats all past gradients equally, and it is thus natural to use exponentially decaying weights for the past gradients. This new definition of $v_t$ leads to another algorithm RMSProp \cite{tieleman2012lecture} (and a more complicated algorithm AdaDelta \cite{zeiler2012adadelta}; for simplicity, we only discuss RMSProp). 
More specifically, at the $t$-th iteration of RMSProp, we randomly pick $i $ and compute $g_t = \nabla F_i ( \theta_t ) $, and then update the second order momentum $v_t$ and parameter $\theta_t$ as
\begin{equation}\label{AdaGrad}
\begin{split}
v_{t} = \beta  v_{t-1} +   ( 1 - \beta )  g_t \circ g_t,    \\
\theta_{t+1} = \theta_t -  \alpha_t   v_t^{-1/2} \circ g_t .
\end{split}
\end{equation}

Adam \cite{kingma2014adam} is the combination of RMSProp and  the momentum method (i.e. heavy ball method). 
At the $t$-th iteration of RMSProp, we randomly pick $i $ and compute $g_t = \nabla F_i ( \theta_t ) $, and then update the first order momentum $m_t$,  the second order momentum $v_t$ and parameter $\theta_t$ as
\begin{equation}\label{AdaGrad}
\begin{split}
 m_t   & = \beta_1  v_{t-1} +   ( 1 - \beta_1 )  g_t ,   \\
v_{t}  & = \beta_2  v_{t-1} +   ( 1 - \beta_2 )  g_t \circ g_t,    \\
\theta_{t+1} & = \theta_t -  \alpha_t   v_t^{-1/2} \circ m_t .
\end{split}
\end{equation}

There are a few other related methods in the area, e.g. AdaDelta \cite{zeiler2012adadelta}, Nadam  \cite{dozat2016incorporating}, and interested readers can refer to \cite{ruder2016overview} for more details. 

 \textbf{Empirical use of adaptive gradient methods.}
 AdaGrad was designed to deal with sparse and highly unbalanced data.  
 Imagine we form a data matrix with the data samples being the columns, then in many machine learning applications,
 most rows are sparse (infrequent features) and some rows are dense (frequent features).
   If we use the same learning rate for all coordinates, then
 the infrequent coordinates will be updated too slowly compared to frequent coordinates. This is the motivation to use different learning rates for different coordinates. 
  AdaGrad was later used in many machine learning tasks with sparse data such as language models where the words have a wide range of frequencies \cite{mikolov2013efficient, pennington2014glove}.

Adam is one of the most popular methods for neural network training nowadays \footnote{The paper that proposed Adam \cite{kingma2014adam}  achieved phenomenal success at least in terms of popularity.  It was posted in arxiv on December 2014;
by Aug 2019, the number of citations
	in Google scholar is 26000; by Dec 2019, the number is 33000. Of course  the contribution to optimization area  cannot just be judged by the number of citations, but the attention Adam received is still quite remarkable. }. 
After Adam was proposed, the common conception was that Adam converges faster than vanilla SGD and SGD with momentum,
 but generalizes worse. Later, researchers found that (e.g., \cite{wilson2017marginal}) well-tuned SGD and SGD with momentum outperform Adam in both training error and test error. 
 Thus the advantages of Adam, compared to SGD, are considered to be the relative insensitivity to hyperparameters and rapid initial progress in training (see, e.g. \cite{keskar2017improving}). 
 \citet{sivaprasad2019tunability} proposed a metric of ``tunability'' and verified that Adam is the most tunable
 for most problems they tested. 
 
  The claim of the ``marginal value'' of adaptive gradient methods \cite{wilson2017marginal} in year 2017 did not stop the booming of Adam in the next two years. Less tuning is one reason, but we suspect that another reason is that the simulations done in \cite{wilson2017marginal} are limited to image classification,
 and do not reflect the real application domains of Adam such as GANs and reinforcement 
 learning.\ifarxiv 
  \footnote{ For the 8 most cited papers in Google Scholar
  	among those citing the original Adam paper \cite{kingma2014adam}, and found that four papers are on GANs (generative adversarial networks) \cite{radford2015unsupervised, isola2017image, ledig2017photo, arjovsky2017wasserstein},   two on deep reinforcement learning \cite{mnih2016asynchronous, lillicrap2015continuous} and two on language-related tasks \cite{xu2015show, vaswani2017attention}. 
  This finding is consistent with the claim in \cite{wilson2017marginal} that ``adaptive gradient methods are particularly popular for training GANs and Q-learning ...''.} 
  \fi 
For these tasks, the generalization ability of Adam  might be a less critical issue. 

\textbf{Theoretical results on adaptive gradient methods.} 
Do these adaptive gradient methods converge?  
Although Adam is known to be convergent in practice and the original Adam paper \cite{kingma2014adam} claimed a convergence proof, it was recently found in \citet{reddi2018convergence} that RMSProp and Adam can be divergent (and thus there is some error in the proof of \cite{kingma2014adam}) even  for solving convex problems.   
To understand the reason of divergence, recall that SGD with constant stepsize $\alpha_t$ 
may not converge \cite{luo1991convergence},  but SGD with diminshing step-size (satisfying a few requirements) converges. 
In AdaGrad,  the ``effective'' stepsize  $ \alpha_t /\sqrt{v_t}  $ is diminishing and AdaGrad converges, but in Adam and RMSProp the effective stepsize $ \alpha_t /\sqrt{v_t}  $ is not necessarily diminishing (even if the step-size $ \alpha_t $ is decreasing), thus causing divergence. 
To fix the divergence issue, \cite{reddi2018convergence} proposed AMSGrad, which changes the update of $v_t$ in Adam to the following:
$$
  \bar{v}_t = \beta_2 \bar{v}_{t-1} + (1 - \beta_2 ) g_t^2,  \quad v_t = \max \{ v_{t-1} ,  \bar{v}_{t}    \}. 
$$
They also prove the convergence of AMSGrad for convex problems  (for diminishing $\beta_1 $). 
Empirically, AMSGrad is reported to have somewhat similar (or slightly worse) performance to Adam.

The convergence analysis and iteration complexity analysis of adaptive gradient methods are established for non-convex optimization problems in a few subsequent works \cite{chen2018convergence, zhou2018convergence, zou2018convergence, de2018convergence, zou2018sufficient, ward2018adagrad}. For example, \cite{chen2018convergence} considers a general Adam-type methods where $v_t$ can be any function of past gradients $g_1, \dots, g_t$ and  establishes a few verifiable  conditions that guarantee the convergence  for non-convex problems (with Lipschitz gradient). 
We refer interested readers to  \citet{barakat2019convergence} which provided a table
summarizing the assumptions and conclusions for adaptive gradient methods. 
Despite the extensive research, 
 there are still many mysteries about adaptive gradient methods.
 For instance, why  it works so well in practice is still largely unknown.


  \subsection{Large-scale distributed computation}
 An important topic in neural network optimization
 is how to accelerate training  by using multiple machines.
 This topic is closely related to distributed and parallel computation (e.g. \cite{bertsekas1989parallel}). 
 
 \textbf{Basic analysis of scaling efficiency.}
 Intuitively, having $K$ machines can speed up training by up to $K $ times.
  In practice, the acceleration ratio depends on at least three factors:
     communication time, synchronization time and convergence speed. 
     Ignoring the communication time and synchronization time, the acceleration ratio of $K $ can be achieved in an extreme case that data on different machines do not share common features. In another extreme case where data on different machines are the same,
     the acceleration ratio is at most $ 1 $.   
     In practice, the acceleration ratio often lies in the region $[ 1 , K ] $. 
    Deep learning researchers often use 
    ``scaling efficiency'' to denote the ratio between the acceleration ratio and the number
    of machines. For instance, if $ K $ machines are used and the multi-machine training is
    $ K/2$ times faster than single-machine training, then the scaling efficiency is $ 0 .5$.  The goal
    is to achieve a scaling efficiency as close to $1$ as possible without sacrificing
     the test accuracy. 

 \textbf{Training ImageNet in 1 hour.}
  \citet{goyal2017accurate}  successfully trained ResNet50 (50-layer ResNet) for the ImageNet dataset in 1 hour using 256 GPUs;  in contrast, the original implementation in \citet{he2016deep} takes 29 hours using 8  GPUs. 
  The scaling efficiency is $ 29/32 \approx 0.906, $ which is remarkable. 
 \citet{goyal2017accurate} used 8192 samples in one mini-batch, while  \citet{he2016deep} only used 256 samples in one mini-batch. 
 Bad generalization was considered to be a major issue for
 large mini-batches, but \cite{goyal2017accurate} argued that  optimization difficulty is the major issue.  They used two major optimization tricks:
   first, they scale the learning rate with the size of the mini-batches;
   second, they use ``gradual warmup'' strategy that increases
   the learning rate from $\eta/K$ gradually to $ \eta $ in the first 5 epochs,
   where $K$ is the number of machines. 
   
   \textbf{Training ImageNet in minutes.}
   Following \citet{goyal2017accurate}, a number of works 
   \cite{smith2018do,akiba2017extremely,jia2018highly,mikami2018massively,ying2018image,yamazaki2019yet} have further reduced the total training time by using more machines.
  For example, \citet{you2018imagenet} applied layer-wise adaptive rate scheduling (LARS)
    to train ImageNet with mini-batch size 32,000  in 14 minutes.
    \citet{yamazaki2019yet} used warmup and LARS, tried many learning rate decay rules
    and used label smoothing to train ImageNet in 1.2 minutes by 2048 V100 GPUs,
    with mini-batch size 81920. 
    Note that all these works train ResNet50 on ImageNet to get validation accuracy
    between 75$\%$ to 77$\%$. 
   Multi-machine computation has also been studied  on other tasks. 
   For instance, \citet{goyal2017accurate} also tested Mask R-CNN for object detection,
   and \citet{you2019reducing} studied BERT for language pre-training.

  \subsection{Other Algorithms}\label{subsec: Other algorithms}
 
\textbf{Other learning rate schedules.}
We have discussed cyclical learning rate and adaptive learning rate. 
Adaptive stepsize or tuning-free step-size has been extensively studied in non-linear optimization
area (see, e.g. \citet{yuan2008step} for an overview).
One of the representative methods is Barzilai-Borwein (BB) method proposed in year 1988  \cite{barzilai1988two}. 
Interestingly, in machine learning area, an algorithm similar to BB method was  proposed in
the same year 1988 in Becker et al.  \cite{becker1988improving} (and further developed in  Bordes et al. \cite{bordes2009sgd}).  
This is not just a coincidence: it reflects the fact that the problems neural-net researchers have been thinking are very similar to those of non-linear optimizers. 
  LeCun et al. \cite{lecun2012efficient} provided a good overview of the tricks for training SGD, especially step-size tuning based on the Hessian information.
  Other recent works on tuning-free SGD  include Schaul \cite{schaul2013no}, Tan et al. \cite{tan2016barzilai} and Orabona \cite{orabona2017training}.

\textbf{Second order methods}.
Second-order methods   
have also been extensively studied in the neural network area.
Along the line of classical second-order methods, Martens \cite{martens2010deep} presented
Hessian-free optimization algorithms, which are a class of quasi-Newton methods without 
explicit  computation of an approximation of the Hessian matrix (thus called ``Hessian free'').
One of the key tricks, based on \cite{pearlmutter1994fast,schraudolph2002fast},  is how to compute  Hessian-vector products efficiently by backpropagation, without computing the full Hessian. 
Berahas \cite{berahas2019quasi} proposed a  stochastic quasi-Newton method 
for solving neural network problems. 
Another tye of second order method  is the natural gradient method \cite{amari2000adaptive, martens2014new}, which scales the gradient  by the empirical Fisher information matrix (based on theory of  information geometry \cite{amari2007methods}).
We refer the readers to \cite{martens2014new} for a nice interpretation of natural gradient method and the survey \cite{bottou2018optimization} for a detailed introduction. 
A more efficient version K-FAC, based on block-diagonal approximation and Kronecker factorization,  is proposed in Martens and Grosse  \cite{martens2015optimizing}.

\textbf{Competition between second order methods and first-order methods.}
Adaptive gradient methods actually use second-order information implicitly, and may
be characterized as second-order method as well (e.g. in \citet{bottou2018optimization}).
Here we still view adaptive gradient methods  as first order methods since they
only use a diagonal approximation of the Hessian matrix; 
in contrast, second order methods use a matrix approximation of the Hessian
in a certain way.
Note that there can be a continuous transition between first and second order methods,
dependent on how much second-order information is used. 

During the early times when neural-nets did not achieve  good performance,
some researchers thought that it is due to the limitation of first-order methods and
it may be crucial to develop fast second-order methods.
In the recent decade,  the trend has reversed and first-order methods have been dominant.   \citet{bottou2008tradeoffs} provides some theoretical justification why first-order methods are enough for large-scale machine learning problems. 
Nowadays some researchers thought  second order methods cannot compete with first-order methods since they may overfit, and SGD has some implicit regularization effect for achieving
good test performance. 
Nevertheless, very recently, second order methods showed some promise:  \citet{osawa2018second} has achieved good test performance on ImageNet using K-FAC (only takes 35 epochs to achieve $75\%$ top-1 accuracy on ImageNet).
It is interesting to see whether second order methods can revive in the future.

\section{ Global Optimization of Neural Networks (GON)}\label{sec: global optimization}

One of the major challenges for  neural network optimization is non-convexity. A general non-convex optimization problem can be very difficult to solve 
   due to sub-optimal local minima. 
The recent success of neural networks suggest that neural-net optimization is far from a worst-case non-convex problem, and finding a global minimum is not a surprise in deep learning noways. 
There is a growing list of literature devoted to understanding this problem. 
For simplicity of presentation, we call this subarea ``global optimization of neural networks'' (GON) 
 \footnote{It is not clear how we should call this subarea. Many researchers use ``(provable) non-convex optimization'' to distinguish these research from convex optimization. However, this name may be confused with the studies of non-convex optimization that focus on the convergence to stationary points.  
 	The name  ``global optimization'' might be confused with research on heuristic methods, while GON is mainly theoretical. Anyhow, let's call it global optimization of neural-nets in this article.}.  We remark that research in GON was partially reviewed in \citet{vidal2017mathematics}, but most of the works we reviewed here appear after \cite{vidal2017mathematics}. 
 
The previous two sections  mainly focus on ``local issues'' of training.
Section \ref{sec: gradient explosion and vanishing} discussed gradient explosion/vanishing, and resolving this issue can ensure the algorithm can move
  locally. Section \ref{sec: gradient explosion and vanishing} discussed
  the convergence speed, but the limitation is that the results only show convergence to local minima (or stationary points).
 In this section, we adopt a global view of the optimization landscape.
 Typical questions include but are not limited to: When can an algorithm converge to global minima?   Are there sub-optimal local minima?  How to pick an initial point that ensures convergence to global minima?   What properties do the optimization landscape have?

\subsection{Related areas}

Before discussing neural networks, we  discuss a few related subareas. 

\textbf{Tractable problems}.
Understanding the boundary between ``tractable'' and ``intractable'' problems has been one of the major themes of optimization area.
The most well-known boundary is probably between convex and non-convex problems.
However, this boundary is vague since it is also known that many non-convex optimization problems can be reformulated as a convex problem (e.g. semi-definite programming and geometric programming). 
We guess that some neural-net  problems are in the class of ``tractable'' problems, though
the meaning of tractability is not clear.   Studying neural networks, in this sense, is not much different in essence from the previous studies of semi-definite programming (SDP), except that a theoretical framework as complete as SDP  has not been developed yet. 

\textbf{Global optimization}.
Another related area is  ``global optimization'', a subarea of optimization
 which aims to design and analyze algorithms that find globally optimal solutions.
The topics include global search algorithms for general non-convex problems (e.g. simulated annealing and evolutionary methods), algorithms designed for specific non-convex problems (possibly discrete) (e.g. \cite{lu2019sensitive}), as well as analysis of the structure of specific non-convex problems (e.g. \cite{Ferreira2019}).

\textbf{Non-convex matrix/tensor factorization}.
The most related subarea to GON  is ``non-convex optimization for  matrix/tensor factorization'' (see, e.g., \citet{chi2019nonconvex} for a survey), which emerged after 2010  in machine learning and signal processing areas \footnote{Again, it is not clear how to call this subarea. ``Non-convex optimization'' might be a bit confusing to optimizers.}.  This subarea tries to understand why many non-convex matrix/tensor problems  can
be solved to global minima easily.  Most of these problems can be viewed as the extensions of matrix factorization problem
\begin{equation}\label{matrix factorization}
	\min_{X, Y \in \mathbb{R}^{n \times r} } \| M - XY^T \|_F^2 ,
\end{equation}
 including low-rank matrix completion,  phase retrieval, matrix sensing, dictionary learning
 and tensor decomposition.
The matrix factorization problem \eqref{matrix factorization} is closely related to the eigenvalue problem.
Classical linear algebra textbooks explain the tractability of the (original) eigenvalue problem by proving directly the convergence of power method, but it cannot easily explain what happens if a different algorithm is used.
 In contrast, an optimization explanation is that the eigenvalue problem can be solved to global optima because every local-min is a 
 global-min. 
One central theme of this subarea is  to study whether a nice geometrical property still holds for
 a generalization of  \eqref{matrix factorization}.
This is similar to GON area, which essentially tries to understand the structure of deep non-linear neural-nets that also can be viewed as generalization  of \eqref{matrix factorization}.

\ifjournal 
\fi

\subsection{Empirical exploration of landscape}\label{subsubsec: Landscape Analysis}

 We first  discuss some interesting empirical studies on the optimization landscape of neural networks. 
 Some of the empirical studies like lottery ticket hypothesis have sparked a lot of interests from practitioners as 
 they see potential practical use of landscape studies. 
 Theoretical results will be reviewed mainly in later subsections.

 One of the early papers that caught much attention is Dauphin et al. \cite{dauphin2014identifying}, which showed
 that empirically bad local minima are not found and  a bigger challenge is plateaus.
 Goodfellow et al. \cite{goodfellow2014qualitatively} plotted the function values
 along the line segment between the initial point and the converged point, and found
 that this 1-dimensional plot is similar to a 1-dimensional convex plot which has no bumps.
 These  early experiments  indicated that the landscape of a neural-net problem  is much nicer than one thought.
  
 A few later works provided various ways to explore the landscape. 
 \citet{poggio2017theory} gave experiments on the visualization of the evolution of SGD. 
 Li et al. \cite{li2018visualizing}  provided visualization of the landscape
 under different network architecture. 
 Baity-Jesi et al.  \cite{baity2018comparing} compared the learning  dynamics of neural-nets with glassy systems  in statistical physics.  Franz et al.  \cite{franz2018jamming} and Geiger et al.  \cite{geiger2018jamming}  studied the analogy between the landscape
 of neural networks and  the jamming  transition in physics.  
 
   \subsubsection{Mode connectivity}
  An exact characterization of a high-dimensional surface  is almost impossible,
   thus goemeters strive to identify simple yet non-trivial properties (e.g.
   Gauss's curvature).
   One such property called ``mode connectivity'' has been found for deep neural networks.  
  In particular, Draxler et al.  \cite{draxler2018essentially}  and  \citet{garipov2018loss} 
  independently found that two global minima can be connected by an (almost) equal-value path.
This is an empirical claim, and in practice the two ``global minima'' refer to  two low-error solutions found by training from two random initial points.  

A more general optimization property is ``connectivity of sub-level sets''.
If the sub-level set $ \{ \theta:   F(\theta) \leq  c   \} $ is connected for $c $ being
the global minimal value, then any two global minima can be connected via
an equal-value path. 
    The connectivity of the sub-level sets was first proved by \cite{freeman2016topology}
  for 1-hidden layer linear networks, and \cite{freeman2016topology}
  also empirically  verified the connectivity for MINST dataset. 
 The contributions of  \citet{draxler2018essentially}  and  \citet{garipov2018loss} 
  are that they used stronger path-finding algorithms to validate the connectivity
    of global minima for CIFAR10 and CIFAR100 datasets. 
  The connectivity for deep neural networks was theoretically justified  in \citet{nguyen2019connected, kuditipudi2019explaining}.

  \subsubsection{Model compression and lottery ticket hypothesis}
 Another line of research closely related to the landscape  is  training
 smaller neural networks (or called ``efficient deep learning'').
 This line of research has a close relation with GON, 
 and this relation  has been largely ignored by both theoreticians and practitioners.

The current neural network models often contain a huge number of parameters (millions or even hundreds of millions).
Models for solving ImageNet classification are already large, and recent models
for other tasks are even bigger (e.g. BERT \cite{devlin2018bert}
and bigGAN \cite{brock2018large}). 
 While understanding the benefit of over-parameterzation has been a hot topic (reviewed later),
 for practitioners it is more pressing  to design new methods to train smaller models.
 Smaller models can be used for resource-constrained hardware (e.g. mobile
 devices, internet-of-things devices), and also accessible to more researchers. 
  However, typically a much smaller models will lead to significantly worse performance.

Network pruning  \cite{han2015deep}  showed that 
many large networks can be pruned to obtain
a much smaller network while the test accuracy is only dropped little.
Nevertheless, in network pruning,  the small network often has to inherit the weights from the solution found by training the large network to achieve good performance, and training a small network from the scratch often leads to significantly worse performance \footnote{There are some recent pruned networks that can be trained from random initial point \cite{liu2018rethinking,lee2018snip}, but the sparsity level is not very high; see \cite[Appendix A]{frankle2019lottery} for discussions. }.

 \citet{frankle2018lottery} made an interesting finding that in some cases  a good initial
point is relatively easy to find. 
More specifically, for some datasets (e.g. CIFAR10),  \cite{frankle2018lottery} empirically
shows that a large network contains a small subnetwork and a certain ``half-random'' initial point such that the following holds:  training the small network from this initial point can achieve  performance similar to the large network. 
\ifarxiv
This ``semi-random'' initial point is found by the following procedure:
 first, record the random initial point $\theta^0 $ for the large network, and train the large network  to converge to get $\theta^*$; second, define a mask
  $ \Omega \in \{ 0 , 1 \}^{ |\theta| } $ as $\Omega(k) = 1$ if $ | \theta^*_k  |  > \delta $ and $\Omega(k) = 0$ if $
   | \theta^*_k |  > \delta $, where $\delta$ is a certain threshold and $\theta^*_k$ denotes
   the $k$-th element of $\theta^*$;
   third, define the new initial point as $\tilde{ \theta}^0 = \Omega \circ \theta^0 $,
   and the new small network by discarding those weights with zero values in $\Omega$.
 In short, the new initial point inherits ``random'' weights from the
 original random initial  point, but it only keeps a subset of the weights and thus
 the remaining weights are not independent anymore. 
 \fi 
 The trainable subnetwork (the architecture and the associated initial point together)
 is called a ``winning ticket'', since it has won an ``initialization lottery''. 
Lottery ticket hypothesis (LTH) states that such a winning ticket always exists. 
Later work  \cite{frankle2019lottery} shows that for larger datasets such as ImageNet,  the
procedure in \cite{frankle2018lottery} needs to be modified to find a good initial point. 
\citet{zhou2019deconstructing} further studies the factors that lead to the success
of the lottery tickets (e.g. they find the signs of the weights are very important). 
For more discussions on LTH, see Section 3.1 of \cite{morcos2019one}. 

The works on network pruning and LTH are mostly empirical, and a clean message is yet to be stated due to the complication of experiments. 
It is an interesting challenge to formally state and theoretically analyze the properties related to  model compression and LTH.

   \subsubsection{Generalization and landscape}
\begin{figure}[H]
 	\centering
 	\subfigure[Wide minima generalize better \cite{keskar2016large}]{
 		\label{Fig3a}
 		\includegraphics[width=0.5\textwidth]{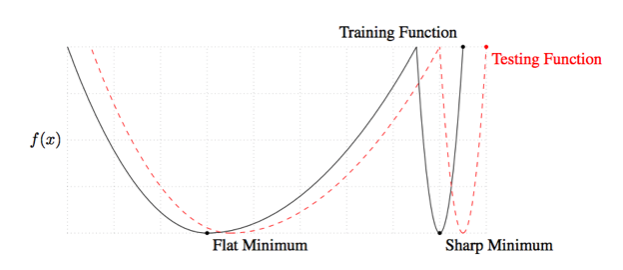}}
 	\subfigure[Sharp minima may become wide after re-parameterization \cite{dinh2017sharp}]{
 		\label{Fig3b}
 		\includegraphics[width=0.3\textwidth]{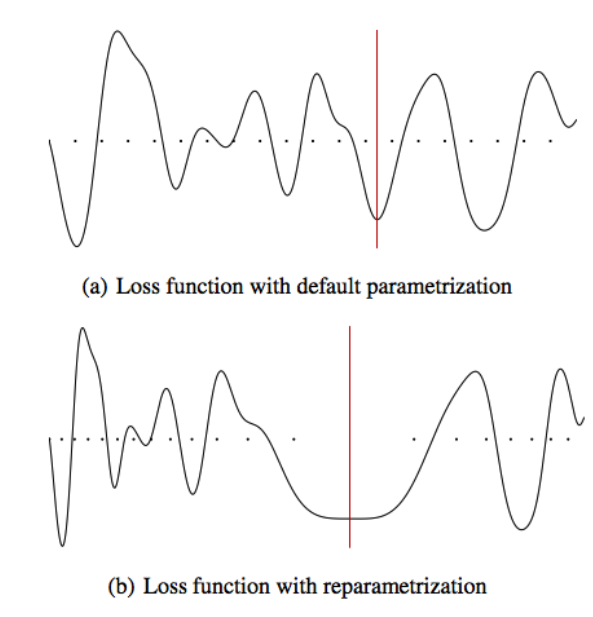}}\hfill 
 	\caption{Illustration on wide minima and sharp minima.}
 	\label{Fig.lable}
 \end{figure}

 Landscape has long been considered to be related to the generalization error.
 A common conjecture is that flat and wide minima generalize better than sharp minima, with numerical evidence in, e.g., Hochreiter and Schmidhuber \cite{hochreiter1997flat}  and Keskar et al. \cite{keskar2016large}. The intuition is illustrated in Figure \ref{Fig3a}:
 the test loss function and  the training loss function have a small difference,  and that
 difference has a small effect on wide minima and thus they generalize well; in constrast,
 this small difference has a large effect on sharp minima and thus they do not generalize well.
 \citet{dinh2017sharp} argues that sharp minima can also generalize since they can become wide minima    after re-parameterization; see Figure \ref{Fig3b}.
   How to define ``wide'' and ``sharp'' in a rigorous way is still challenging. 
   \citet{neyshabur2015path,yi2019positively} defined new metrics for the ``flatness''
   and showed the connection between generalization error and the new notions
   of ``flatness''. 
   \citet{he2019asymmetric} found that besides wide and shallow local minima, there are asymmetric minima
   that the function value changes rapidly along some direction  and slowly along some other directions,
   and algorithms biased towards the wide side generalize better.

Although the intuition ``wide minima generalize better'' is debatable,  
 researchers still borrow this intuition to design or discuss optimization algorithms. 
 \citet{chaudhari2016entropy} designed entropy-SGD that explicitly search for wider minima. 
 \citet{smith2017super} also argued that the benefit of cyclical learning rate  is that it can escape shallow local minima

  \subsection{Optimization Theory  for Deep Neural Networks}\label{subsubsec: Landscape Analysis}
 We discuss two recent threads in optimization theory for \textit{deep} neural networks:
landscape analysis and gradient dynamics analysis. 
 The first thread discusses the global landscape properties  of the loss surface,
 and the second thread studies  gradient dynamics of ultra-wide networks.

 \subsubsection{Global landscape analysis of deep networks}\label{landscape analysis}
 
 Global landscape analysis is the closest in spirit to the empirical explorations in Section \ref{subsubsec: Landscape Analysis}: understanding some geometrical properties the landscape.
 There are three types of   deep neural networks with positive results so far:
  linear networks, over-parameterized networks and modified networks. 
  We will also discuss some negative results.    

\textbf{Deep linear networks.}
Linear networks  have little representation power and are not very interesting from a learning perspective, 
but it is a valid problem from optimization perspective.
The landscape of deep linear networks are relatively well understood. 
 Choromanska et al. \cite{choromanska2015loss}   uses spin glass theory to analyze deep 
linear neural-nets (started from ReLU network, but actually analyzed linear network), and proved that local minima have highest chance to be close to global minima among all stationary points (the precise statement is very technical). 
 Kawaguchi \cite{kawaguchi2016deep} proves that for a deep fully-connected  linear network with quadratic loss, under mild conditions (certain data matrices are full-rank and output dimension $d_y$ is no more than input dimension $d_x$), every local-min is a global-min.
 Lu and Kawaguchi \cite{lu2017depth} provides a much simpler proof for this result under stronger conditions. 
  Laurent and James \cite{laurent2018deep} extends this result to arbitrary loss functions, and Zhang \cite{zhang2019depth} gives a further simplified proof. Hardt and Ma \cite{hardt2016identity} analyzes the number of stationary points in a small region around global minima for linear ResNet.  
Nouiehed and Razaviyayn \cite{nouiehed2018learning} provided a general sufficient condition for the local-min of a neural-net to be global-min, and apply this condition to deep linear networks (also give a weaker result for non-linear pyramid networks).
   Besides characterizing local minima,  stronger claims on the stationary points can be proved for linear networks. 
  Yun et al. \cite{yun2017global} and Zou et al. \cite{zhou2018critical} present necessary and sufficient conditions for a stationary point  to be a global minimum. 

\begin{figure}
 	\centering
 	\includegraphics[width=10cm,height=3cm]{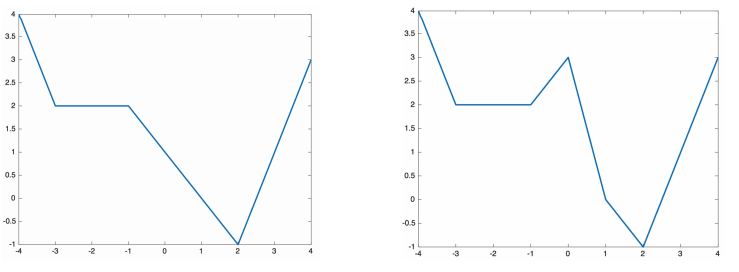}
 	\caption{Left figure: the flat region is not a set-wise strict local-min, and this region can be escaped by a (non-strictly) decreasing algorithm.  Right figure: there is a basin that is a set-wise strict local-min.}
 	\label{fig1}
 \end{figure}

\textbf{Deep over-parameterized networks.}
Over-parameterized networks are the simplest non-linear networks that currently can be analyzed, but already somewhat subtle. 
It is widely believed that ``more parameters than necessary'' can smooth the landscape
\cite{livni2014computational,neyshabur2017exploring,zhang2016understanding}, but these works do not provide a rigorous result.
To obtain rigorous results, one  common assumption for deep networks  is that the last layer has more neurons than the number of samples. 
Under this assumption on the width of the last layer,
Nguyen et al. \cite{nguyen2018loss}  and Li et al. \cite{li2018over}  prove that a fully connected network has no ``spurious valley'' or ``set-wise strict local minima'', under mild assumptions on the data. 
The difference is that Nguyen et al. \cite{nguyen2018loss}  requires  the activation functions satisfy some conditions (e.g. strictly increasing) and can extend the result to other connection patterns (including CNN), and  Li et al. \cite{li2018over} 
 only requires the activation functions to be continuous (thus including ReLU
 and swish). 
  Intuitively,  ``set-wise strict local minima'' and ``spurious valley''
    are the ``bad basin'' illustrated in the right figure of 
   Figure \ref{fig1} (see \cite{nguyen2018loss} or \cite{li2018over} for formal definitions). 

The above works are the extensions of a classical work \cite{yu95} on 1-hidden-layer over-parameterized networks (with sigmoid activations), which claimed to have proved that every local-min is a global-min. 
It was later found in \cite{li2018over} that the proof is not rigorous.
\citet{ding2019spurious} further constructs sub-optimal local-min for arbitrarily wide neural networks
for a large class of activations including sigmoid activations, thus 
under the settings of \cite{yu95}\cite{nguyen2018loss} \cite{li2018over}  sub-optimal local minima
can exist.   This implies that overparameterization cannot eliminate bad local minima, but only
   bad basins (or spurious valleys). 

Finally, it seems that over-parameterized networks are prone to over-fitting, but many practical networks are indeed over-parameterized and understanding why over-fitting does not happen is an interesting line of research \cite{neyshabur2017exploring,bartlett2017spectrally,wei2018margin, wu2017towards,belkin2018reconciling,mei2019generalization}.  In this article,
we mainly discuss the research on the optimization side.

 \textbf{Modified problems.}
 The results discussed so far mainly study the original neural network problem \eqref{main problem}, 
  and the landscape is different if the problem is slightly changed. 
Liang et al. \cite{liang2018adding} provides two modifications, each of which can ensure no bad local-min exists, for binary classification. 
Kawaguchi et al. \cite{kawaguchi2019elimination} extends the result of \cite{liang2018adding} to multi-class classification problems. In addition, \cite{kawaguchi2019elimination}  provides toy examples to illustrate the limitation of only considering local minima: GD may diverge for the modified problem. 
It is  a possible weakness of any result on ``no bad local-min'' including
the classical works on deep linear networks. 
In fact,  as discussed in Section \ref{subsec: convergence anlaysis of GD}, the possibility of divergence (U3) is one of the three undesirable situations  that classical  results on GD does not exclude, and eliminating bad local-min only excludes (U1).

\textbf{Negative results.}
Most of the works in GON area after 2012  are  positive results.
However, while neural-nets can be trained in some cases with careful choices of architecture, initial points and parameters,  there are still many cases that neural-nets cannot be
successfully trained. 
Shalev et al. \cite{shalev2017failures} explained a few possible reasons of failure of GD for training neural networks. 
There are a number of recent works focusing the existence of bad local minima. 

These negative results differ by their assumptions on activation functions, data distribution and network structure. 
As for the activation functions,  many works showed that ReLU networks have bad local minima (e.g., 
Swirszcz et al.\cite{swirszcz2016local}  Zhou et al. \cite{zhou2017critical}, Safran et al.\cite{safran2017spurious}, Venturi et al.\cite{venturi2018spurious}, Liang et al.\cite{liang2018understanding}), and a few works  Liang et al. \cite{liang2018understanding},
Yun et al.\cite{yun2018small}  and \citet{ding2019spurious}  construct examples for smooth activations.
As for the loss function,  \citet{safran2017spurious} and \citet{venturi2018spurious} analyze the population risk (expected loss) and other works analyze the empirical risk (finite sum loss).
As for the data distribution, most works consider data points that lie in  a zero-measure space
or satisfy special requirements like linear separability (\citet{liang2018understanding}) or 
Gaussian (Safran et al.\cite{safran2017spurious}), and few consider generic input data (e.g. \citet{ding2019spurious}). 
We refer the readers to \citet{ding2019spurious} which compared various counter-examples
in a table.

\subsubsection{ Algorithmic analysis of deep networks}

A good landscape may make an algorithm easier to find global minima, but does not fully explain
the behavior of specific algorithms. 
To understand specific algorithms, convergence analysis is more desirable. However, for a general
neural-net the convergence analysis is extremely difficult,  
thus some assumptions have to be made.
The current local (algorithmic) analysis of deep neural-nets  is mainly performed for two types:
linear networks \cite{saxe2013exact, bartlett2018gradient, arora2018convergence, ji2018gradient}  and ultra-wide networks.

\textbf{Linear networks.}
As discussed earlier, gradient explosion/vanishing can cause great difficulty of training neural-nets, and even for the scalar problem $\min_{w_1, \dots, w_L} (1 - w_1 \dots w_L)^2$, it takes GD exponential time to converge \cite{shamir2018exponential}.   
Perhaps a bit surprisingly, for deep linear networks in higher dimension, polynomial time convergence can still be established. 
Arora et al. \cite{arora2018convergence} considered the problem $ \min_{ W_1, \dots, W_L } \| W_1  W_2 \dots W_L  - \Phi \|_F^2 $, and prove that if the initial weights are ``balanced'' and the initial product $W_1 \dots W_L $ is close to $\Phi$, GD with a small stepsize converges to global minima in polynomial time.  Ji and Telgarsky \cite{ji2018gradient} assume linearly separable data and prove that if the initial objective value is less than a certain threshold, then GD with small adaptive stepsize converges asymptotically to global minima. Moreover, they proved that the normalized weight matrices  converge to rank-1 matrices, which matches the empirical observation that
the converged weight matrices are approximately low rank in AlexNet. 
The strong assumptions of these works on  initialization, small stepsize  and/or data are still far from satisfactory, but at least some of these assumptions are necessary in the worst-case (as discussed in \cite{arora2018convergence}).
\citet{shin2019effects} analyzed layerwise-training for deep linear networks, and showed that under some conditions, gradient descent convergences faster for deeper networks.

\textbf{Neural Tangent Kernal (NTK)}. 
Convergence analysis for deep non-linear networks is much harder than linear networks, even under the extra assumption of over-parametrization. Some progress has been made recently. We first discuss the result of \citet{jacot2018neural}  on NTK. 

This NTK result is an extension of a property of linear regression.
A typical explanation why GD converges to global minima of linear regression is that the objective function is convex,
then for neural networks one would extend convexity to other geometrical properties
(basically the idea behind landscape analysis).
There is another explanation from the perspective of gradient flow.
Consider the linear regression problem $ \min_{w \in \mathbb{R}^d  } F(w) \triangleq  \frac{1}{2} \sum_{i = 1}^n (  w^T x_i - y_i  )^2   $. The gradient flow is
  $   \frac{d w(t) }{d t} = - XX^T w(t) +  X y     $, where $X = [x_1, \dots, x_n] \in \mathbb{R}^{d \times n}$.
Define  $ r_i  = w^T x_i - y_i , i =1, \dots, n $, then the dynamics
of the residual $r= ( r_1 ;  \dots; r_n) \in \mathbb{R}^{n \times 1 }$  is  
\begin{equation}\label{dynamics of linear regression}
  \frac{d r (t) }{ dt } = - X^T X r (t). 
\end{equation}
This is called \textit{kernel gradient descent} with respect to the kernel $K = X^T X  \succeq  0 $. 

Consider the neural-network problem with quadratic loss $ \min_{\theta } \sum_{i=1}^n \frac{1}{2} ( f_{\theta}(x_i) - y_i )^2    $, where $x_i \in \mathbb{R}^d,  y_i \in \mathbb{R}$
 (it can be generalized to multi-dimensional output and non-quadratic loss). 
The gradient descent dynamics is  
\begin{equation}\label{dynamics of GD}
	 \frac{d \theta }{ d t } =  
	- \sum_i  \frac{  \partial  f_{\theta }(x_i )   } { \partial  \theta  } ( f_{\theta} (x_i) - y_i  )  .
\end{equation}
Define $ G = ( \frac{ \partial f_{\theta}(x_1)  }{ \partial \theta }, \dots,  \frac{ \partial f_{\theta}(x_n)  }{ \partial \theta } )  \in \mathbb{R}^{ P \times n } $ where $P$ is the number of parameters,
and define  \textit{neural tangent kernel}  $K = G^T G  $. 
Let $ r = ( f_{\theta}(x_1 ) - y_1 ;  \dots ;  f_{\theta}(x_n ) - y_n ) $, then
$ \frac{ d r_i }{ d t}  = \frac{ \partial f_{\theta}(x_i)  }{ \partial \theta } \sum_j  
 \frac{ \partial f_{\theta}(x_j )  }{ \partial \theta } r_j $, or equivalently,
\begin{equation}\label{dynamics of neural-nets}
	 \frac{d r  }{ d t }  = K (t)  r ,
\end{equation}
When $ f_{\theta}(x )  = \theta^T x $, the matrix $ K(t) $ reduces to a constant matrix $X^T X$, thus
\eqref{dynamics of neural-nets} reduces to \eqref{dynamics of linear regression}.

 \citet{jacot2018neural} proved that $ K(t) $ is a constant matrix 
 for any $t $  under certain conditions.   
 More specifically, if the initial weights are i.i.d. Gaussian
with certain variance (similar to LeCun initialization),
then as the number of neurons at each layer goes to infinity sequentially,  
 $ K(t)$ converges to  a constant matrix $K_{c} $ (uniformly for all $t \in [0, T]$ where
$T$ is a given constant). 
Under further assumptions on the activations (non-polynomial activations) and data (distinct data from
the unit sphere),  \cite{jacot2018neural} proves that $K_c $ is positive definite. 
One  interesting part of \cite{jacot2018neural} is 
 that  the limiting NTK matrix $K_c$ has a closed form expression,
computed recursively by an analytical formula. 

\citet{yang2019scaling} and \citet{novak2019bayesian} extended \cite{jacot2018neural}: 
they only require  the width of each layer goes to infinitely simultaneuously (instead of sequentially in \cite{jacot2018neural}), and provides a formula of NTK for convolutional networks,
called CNTK.


\textbf{Finite-width Ultra-wide networks.}
Around the same time as \cite{jacot2018neural},  Allen-Zhu et al. \cite{allen2018convergence} and  Zou et al. \cite{zou2018stochastic}  and Du et al.  \cite{du2018gradient}
 analyzed deep ultra-wide non-linear networks and prove that with Gaussian initialization and small enough step-size, GD and/or SGD converge to global minima
  (these works can be viewed extensions of an analysis of a 1-hidden-layer networks \cite{li2018learning,du2018gradient}).  
In contrast to the landscape results \cite{li2018over,nguyen2018loss} that only require one layer to have $n$ neurons,
these works require a much larger number of neurons per layer: $ O(n^{24} L^{12} /\delta^8 ) $  in  \cite{allen2018convergence} where
$\delta = \min_{i\neq j} \| x_i - x_j \|$ and $O( n^4 /\lambda_{\min }(K)^4  )$ in \cite{du2018gradient} where $K$ is a complicated matrix defined recursively. 
   \citet{arora2019exact} also analyzed finite-width networks, by proving a non-asymptotic version of the NTK result of \cite{jacot2018neural}.
\citet{zhang2019training,ma2019analysis} analyzed the convergence of over-parameterized ResNet. 

\textbf{Empirical computation by NTK.}
The explicit formula of the limiting NTK makes it possible to actually compute NTK and perform kernel gradient descent for a real-world problem. 
As computing  the CNTK directly is time consuming,  \citet{novak2019bayesian} used Monte Carlo sampling to approximately
compute CNTK. 
\citet{arora2019exact} proposed  an exact efficient algorithm to compute CNTK and tests it on CIFAR10, achieving $ 77 \%  $ test accuracy for CNTK with global average pooling.
\citet{li2019enhanced} utilized two further tricks to achieve 89$\%$ test accuracy on CIFAR10, on par with AlexNet.

\textbf{Mean-field approximation: another group of works.}
There are another group of works which also studied infinite-width limit of SGD. 
  \citet{sirignano2019mean} considered discrete-time SGD for infinite-width multi-layer neural networks, and showed that the limit of the neural network output satisfies
  a certain differential equation. 
  \citet{arajo2019meanfield,nguyen2019mean} also studied infinite-width multi-layer networks.
These works are extensions of previous works 
Mei et al. \cite{mei2018mean}, Srignanao and Spiliopoulos \cite{sirignano2018mean} and Rotskoff and Vanden-Eijnden \cite{rotskoff2018neural},
which  analyzed 1-hidden-layer networks. 
A major difference between these works and \cite{jacot2018neural} \cite{allen2018convergence} \cite{zou2018stochastic} \cite{du2018gradient} is the scaling factor;
for instance, \citet{sirignano2018mean} considered the scaling factor $1/\text{fan-in}$, while \cite{jacot2018neural} \cite{allen2018convergence} \cite{zou2018stochastic} \cite{du2018gradient} considered the scaling factor $1/ \sqrt{\text{fan-in}} $.
 The latter scaling factor of $1/ \sqrt{\text{fan-in}} $ is used in LeCun initialization (corresponding
 to variance $1/ \text{fan-in} $), thus closer to practice,
 but they imply that the parameters mover very little as the number of parameters increase.
In contrast,  \cite{mei2018mean,sirignano2018mean, rotskoff2018neural,sirignano2019mean,arajo2019meanfield,nguyen2019mean}  show that the parameters evolve according to a PDE and thus can move far away from the initial point.

\textbf{``Lazy training'' and two learning schemes.}
The high-level idea of \cite{jacot2018neural} \cite{allen2018convergence} \cite{zou2018stochastic} \cite{du2018gradient} is  termed ``lazy training'' by \cite{chizat2018lazy}:  the model behaves like its linearization around its initial point.  
Because of the huge number of parameters, each parameter only needs to move 
a tiny amount, thus linearization is a good approximation.
However, practical networks are not ultra-wide, thus the  parameters will move a reasonably 
large amount of distance, and likely to move out of the linearization regimes.
\cite{chizat2018lazy} indeed showed that the behavior of SGD in practical neural-nets is different from lazy training.
In addition, \cite{chizat2018lazy} pointed out that ``lazy training'' is mainly due to implicit
choice of the scaling factor, and applies to a large class of models beyond neural networks. 
A natural question is whether the ``adaptive learning scheme'' described by
 \cite{mei2018mean,sirignano2018mean, rotskoff2018neural,sirignano2019mean,arajo2019meanfield,nguyen2019mean}
 can partially characterize the behavior of SGD.
 In an effort to answer this question,  \citet{williams2019gradient}  analyzed a 1-hidden-layer ReLU network with 1-dimensional input, and provided conditions for the ``kernel learning scheme'' and ``adaptive learning scheme''.

\textbf{Discussions}.
Math is always about simplification.  Landscape analysis ignores the algorithmic aspects and focus on geometry (like geometricians). 
Analysis of gradient dynamics provides a more precise description of the algorithm (like dynamical systems theorists), but requires strong assumptions such as a very large width.
A major difference is the point of departure.
 Landscape analysis only studies one aspect of the whole theory (as discussed 
 in Section \ref{subsec: big picture of decomposition}, this is common in machine learning), 
while algorithmic analysis aims to provide an end-to-end analysis that covers
 all aspects of optimization. 
 From a theoretical perspective, it is very difficult to understand
 every aspect of an algorithm (even for interior point methods there are unknown questions),
 thus some aspects have to be ignored. 
The question is whether essential aspects have been captured and/or ignored. 
 One may argue that the trajectory of the algorithm is crucial, thus landscape analysis 
 ignores some essential part.
 One could also argue that moving outside of a tiny neighborhood is important,
 thus ``lazy training'' ignores some essential part.
Nevertheless, from the angle of extracting some useful insight,
 landscape analysis has led to the discovery of mode connectivity and
 algorithmic analysis has led to empirical CNTK, so both have shown their potential.

\subsection{Research in Shallow Networks  after 2012}

For the ease of presentation, results for shallow networks are mainly reviewed in this subsection. 
Due to the large amount of literature in GON area, it is hard to review all recent works,
and we can only give an incomplete  overview.  
We group these works based on the following criteria: 
landscape or algorithmic analysis (first-level classification criterion);  
one-neuron, 2-layer network or 1-hidden-layer network \footnote{In this section, we will use ``2-layer network'' to denote a network like $ y = \phi ( W x + b  ) $ or $y = V^* \phi (W x + b)$ with fixed $V^*$, and use ``1-hidden-layer network'' to denote a network like $y = V \phi (W x + b_1 ) + b_2 $ with both $V$ and $W$ being variables. }
 (second-level criterion).
Note that among the works in the same class, they may differ on the  assumption on input data (Gaussian input and linearly separable input are common),  number of neurons,  loss function  and specific algorithms (GD, SGD or others). 
Note that this section focuses on positive results, and
negative results for shallow networks are discussed in Section \ref{landscape analysis}. 

  \textbf{Global landscape of 1-hidden-layer neural-nets.}
 There have been many works on the landscape of 1-hidden-layer neural-nets. 
 One interesting work (mentioned earlier when discussing mode connectivity) is Freeman and Bruna \cite{freeman2016topology} which  proved that the sub-level set is connected  for deep linear networks and 1-hidden-layer ultra-wide ReLU networks. This does not imply every local-min is global-min, but implies there is no spurious valley (and no bad strict local-min). A related recent work is Venturi et al. \cite{venturi2018neural}  which proved no spurious valley exists (implying no bad basin) for 1-hidden-layer network with ``low intrinsic dimension''. Haeffele and Vidal \cite{haeffele2017global} extended the classical work of Burer and Monteiro \cite{burer2005local} to 1-hidden-layer neural-net, and proved that a subset of the local minima are global minima,
 for a set of positive homogeneous activations. 
 Ge et al. \cite{ge2017learning} and Gao et al. \cite{gao2018learning} designed a new loss function so that all local minima are global minima. Feizi et al. \cite{feizi2017porcupine} designed a special  network for which almost all local minima are global minima. Panigrahy et al.  \cite{panigrahy2017convergence} analyzed local minima for many special neurons via electrodynamics theory. 
For quadratic activations, Soltanolkotabi et al. \cite{soltanolkotabi2019theoretical} proved that 2-layer over-parameterized network (with Gaussian input) have no bad local-min, and Liang et al. \cite{liang2018understanding} provided a sufficient and necessary condition for the data distribution so that 1-hidden-layer neural-net has no bad local-min. 
For 1-hidden-layer ReLU networks (without bias term), Soudry and Hoffer  \cite{soudry2017exponentially} proved that the number of differentiable local minima is very small. Nevertheless,  Laurent and von Brecht \cite{laurent2017multilinear} showed that except flat bad local minima, all local minima of 1-hidden-layer ReLU networks (with bias term) are non-differentiable.


  \textbf{Algorithmic analysis of 2-layer neural-nets.}
There are many works on the algorithmic analysis of SGD for shallow networks under a variety of settings. 
The first class analyzed SGD for 2-layer neural-networks (with the second layer weights fixed). 
A few works mainly analyzed one single neuron.
Tian \cite{tian2017analytical} and Soltanolkotabi \cite{soltanolkotabi2019theoretical} analyzed the performance of GD
for a single ReLU neuron. 
Mei et al. \cite{mei2018mean} analyzed  a single sigmoid neuron. 
Other works analyzed 2-layer networks with multiple neurons.
Brutzkus and Globerson \cite{brutzkus2017globally} analyzed a non-overlapping 2-layer ReLU network  and proved that the problem is NP-complete for general input, but if the input is Gaussian then GD converges to global minima in polynomial time.
Zhong et al. \cite{zhong2017recovery} analyzed 2-layer under-parameterized network (no more than $d$ neurons)
for Gaussian input and initialization by tensor method. 
Li et al.  \cite{li2017convergence} analyzed 2-layer network with skip connection for Gaussian input.
Brutzkus et al. \cite{brutzkus2017sgd} analyzed 2-layer over-parameterized network with leaky ReLU activation for linearly seperable data. Wang et al. \cite{wang2018learning} and Zhang et al. \cite{zhang2018learning} analyzed  2-layer over-parameterized network with ReLU activation, for linearly separable input and Gaussian input respectively.  Du et al. \cite{du2018power} analyzed 2-layer over-parameterized network with quadratic neuron for Gaussian input. 
Oymak and  Soltanolkotabi \cite{oymak2019towards} proved the global convergence of GD with random initialization for a 2-layer network with a few types of neuron activations, when the number of parameters exceed $O(n^2)$ ($O(\cdot)$ here hides condition number and other parameters). 
\citet{su2019} analyzed GD for 2-layer ReLU network with $O(n)$ neurons for generic input data. 

  \textbf{Algorithmic analysis of 1-hidden-layer neural-nets.}
The second class analyzed 1-hidden-layer neural-network (with the second layer weights trainable). 
  The relation of 1-hidden-layer network and tensors is explored in  \cite{janzamin2015beating, mondelli2018connection}. 
Boob and Lan \cite{boob2017theoretical} analyzed a specially designed alternating minimization method for over-parameterized 1-hidden-layer neural-net.  
Du et al. \cite{du2017gradient} analyzed an non-overlapping network for Gaussian input and with an extra normalization, and proved that SGD can converge to global-min for some initialization and converge to bad local-min for other initialization. 
Vempala and Wilmes \cite{vempala2018polynomial} proved that for random initialization and  with $n^{O(k)}$ neurons,  GD converges to the best degree $k$ polynomial approximation of the target function; a matching lower bound is also proved. 
Ge et al. \cite{ge2018learning} analyzed a new spectral method for learning 1-hidden-layer network. 
Oymak and Soltanolkotabi \cite{oymak2018overparameterized} analyzed GD for a rather general problem and applied it to 1-hidden-layer neural-net where $n \leq d$ (number of samples no more than dimension) for any number of neurons.

\section{Concluding Remarks}

In this article, we have reviewed existing theoretical results related to neural network optimization, mainly 
focusing on the training of feedforward neural networks. 
The goal of theory in general is two-fold: understanding and design. 
As for understanding,  now we have a good understanding on the effect of initialization on stable training, and some understanding  on the effect of over-parameterization on the landscape.
As for design,   theory has already greatly helped  the design of algorithms (e.g. initialization schemes, batch normalization, Adam). 
There are also examples like CNTK that is motivated from theoretical analysis and has become a real tool. 
Besides design and understanding, some interesting empirical phenomenons have been discovered, such as mode connectivity and lottery ticket hypothesis, awaiting more theoretical studies.
Overall, there is quite some progress in the theory for neural-net optimization.

\ifarxiv
That being said, there are still lots  of challenges.     
As for understanding, we still do not understand many of the components that affect the performance, e.g., the detailed architecture and Adam optimizer. 
With the current theory, it is still far from making a good prediction on the performance of an algorithm, especially in a setting that is different from the classification problem. 
As for design, one major challenge for the theoretical researchers is that the chance of (strong) theoretically-driven algorithms for image classification seems low. 
Opportunities  may lie in other areas, such as robustness to adversarial attacks and deep  reinforcement learning. 
\fi 

\section{Acknowledgement}
We  would like to thank Leon Bottou, Yan Dauphin, Yuandong Tian,  Levent Sagun, Lechao Xiao, Tengyu Ma, Jason Lee, Matus Telgarsky, Ju Sun,  Wei Hu, Simon Du,  Lei Wu, 
Quanquan Gu, Justin Sirignano, Shiyu Liang, R. Srikant, Tian Ding and Dawei Li  for discussions on various results reviewed in this article.  We also thank Ju Sun for the list of related works in the webpage \cite{sunBlog} which helps the writing of this article.

\appendix

\section{Discussion of General Convergence Result}\label{appen: convergence discussions}

This section is an extension of  Section \ref{subsec: convergence anlaysis of GD}
on the convergence analysis.

\textbf{Convergence of iterates.}
Recall that the statement ``every limit point is a stationary point'' does not eliminate
two undesirable cases: (U1)  the sequence could have more than one limit points; 
(U2)  limit points could be non-existent.

It is relatively easy to eliminate the possibility of (U1) since for most neural network training problem, Kurdyka-Lojasiewicz condition holds (see, e.g. \cite{zeng2018block}), and together with some minor conditions, it is not hard to show that  there can be no more than one limit points for a descent algorithm  (see, e.g., \cite{luo1993error, attouch2013convergence}). 
Nevertheless, a rigorous argument for a generic neural network optimization
problem is not easy. 

Eliminating the possibility of (U2)  is both easy and hard. 
It is easy in the sense that the divergence is often excluded by adding extra assumptions such as compactness of  level sets $\{ x \mid f(x ) \leq c  \}$, which can be enforced by adding a proper regularizer. It is hard since for neural-net optimization, the required regularizer 
may be impractical (e.g. a degree $2L$ polynomial for quadratic loss). 
Another solution is to add a ball constraint on the variables, but that will cause  other issues (e.g. convergence analysis of SGD for constrained problems is complicated).  Thus,  if one really wants to eliminate (U2), a tailored analysis for neural-nets may be needed.

\textbf{Global Lipschitz constants.}
Global Lipschitz smoothness of gradient is required for GD to converge, but neural network optimization problems do not have a global Lipschitz constant. 
 An intuitive solution is to use a local Lipschitz constant for picking the stepsize  instead of a global Lipschitz constant,  but it seems hard to provide a clean result.
 We discuss a few plausible solutions and mention their issues. 

A natural choice is to use stepsize dependent on local Lipschitz constant (i.e.
the largest eigenvalue of the Hessian at the current iterate). 
To prove the convergence, we can modify the proof of Proposition 1.2.3  in \cite{bertsekas1997nonlinear}, but the proof does not work directly
since the step-size could go to zero too fast, and does not satisfy the condition
$\sum_t \eta_t = \infty  $.  
One may wonder whether GD with stepsize dependent on local Lipschitz constant is a special case of the ``successive upper-bound minimization'' framework of \cite{razaviyayn2013unified} and the convergence theorem such as Theorem 3 in  \cite{razaviyayn2013unified}  can directly apply.
However, it is not a special case of \cite{razaviyayn2013unified} since that paper requires a ``global upper bound'' of the objective function, but using a local Lipschitz constant only provides a local upper bound.

Another idea is to utilize the fact that the gradient is Lipschitz smooth in a compact set such as 
a ball.  For instance, for matrix factorization problems, 
using local Lipschitz constant in a ball is a common approach, e.g.,
Lemma 1 of \cite{chi2019nonconvex}. However,  this lemma also requires convexity in the ball to ensure the iterates do not move out of the ball.  
For general non-convex functions, it is not easy to prove that the algorithm does not move out 
of the ball. As an exercise, readers can try to analyze $ \min_{w} (1 - w^6 )^2  $, 
and see whether it is easy to prove GD with stepsize chosen
based on the Lipschitz constant in a ball converges \footnote{We do not know a clean proof
that is generalizable to high-dimensional problems. All our current proofs utilize
certain property of the problem which highly relies on the 1-dimensional structure,
and are thus not that interesting.}.

To remedy the idea of using Lipschitz constant of a compact set, a natural solution is to 
 keep the iterates bounded. 
But how to ensure the iterates are bounded? One way is to add constraints on the variables, but this may cause other difficulties since it becomes a constrained problem.
Another solution is to add regularizers such as $ \| \theta \|^2 $,
 $ \max \{   \| \theta \|^2,  B     \}  $, 
 or a smooth version $  ( \max \{   \| \theta \|^2 - B  ,  0     \} ) ^2   $
as used in \cite{sun2016guaranteed}, where $B$ is large enough such that at least
one global minimum has norm no more than $ \sqrt{ B} $. 
  However, it  seems not easy to rigorously prove convergence even with
  the aid of regularizers.
   In addition, the existing landscape analysis or global convergence are done for non-regularized problems, and new analysis is required if regularizers are added. 
 Again, analysis tailored for a specific problem may prove these, but for now we are hoping for a clean universal analysis. 
In short,  a simple modification to the problem seems hard to ensure the existence of a constant stepsize that guarantees the convergence \footnote{In fact, even dealing with convex problems without Lipschitz constant is not an easy problem in optimization, and only until recently are there good progress in some convex problems \cite{bauschke2016descent, lu2018relatively}. }. 

For most practitioners, there is a simple conceptual solution to understand convergence: 
adding a ``posterior'' assumption on the boundedness of iterates. More specifically, assuming the iterates are bounded, then GD with a proper constant step-size 
converges in the sense that the gradient converges to $0$.
The assumption itself can be verified in practice, but it is only verifiable after running
the algorithm (thus a ``posterior'' assumption). 
Anyhow, this is one of the many imperfections of the current theory, and we have to put it aside 
and move on to other aspects of the problem.

\section{Details of Batch Normalization}\label{appen: batch norm}

 \textbf{First trick of BatchNorm}. 
There are two extra tricks in the implemented BatchNorm, and the first trick is to add two more parameters to restore the representation power.
We define a formal BN operation as follows.
For scalars $a_1, a_2, \dots, a_N $,   define $\mu = \frac{1}{N}(a_1 + \dots + a_N ) $
and $\sigma =  \sqrt{ \frac{1}{N} ( a_i - \mu )^2 }   $, then
$$ \mathrm{BN}_{ \gamma, \beta }(a_1, \dots, a_N ) \triangleq
\left(  \gamma \frac{a_1 - \mu }{ \sigma + \epsilon } + \beta,   \dots, 
\gamma \frac{a_N - \mu }{ \sigma + \epsilon } + \beta    \right) ,$$
where $\gamma \in \mathbb{R}^+ $ and $ \beta \in \mathbb{R} $  are parameters to be learned,
and $\epsilon$ is a fixed small constant. 
This BN operation is a mapping from input $a_1, \dots, a_N$ 
to output $ \mathrm{BN}_{ \gamma, \beta }(a_1, \dots, a_N )$ and is differentiable. 
BN operation is defined as a general function applicable to any $N$ scalars, and we discuss
how to incorporate it into the neural network next. 
In a neural network, this BN operation is added before each nonlinear transformation layer in the neural network $f_{\theta}$ to obtain a new neural network  $ \tilde{f}_{ \tilde{ \theta}  } $, where $\tilde{ \theta}$ involves the new parameters  $\{ \gamma^l, \beta^l \}_{l=1}^L $.

We illustrate by a 1-dimensional 2-layer neural network.
Suppose the input instances are $x_1, \dots, x_n \in \mathbb{R}$,
and the original neural network is $ \hat{y}_i =  v  \phi(  w x_i  ), i=1, \dots, n $ where $v, w \in \mathbb{R}$. 
The new network with BN operations is $ ( \hat{y}_1, \dots, \hat{y}_n ) =  
v  \phi(  \mathrm{BN}_{\gamma, \beta }(  w  x_1, \dots, w x_n  )   ) $. 
The problem formulation has also been changed: previously, the objective function can be decomposed  across samples $ F(\theta) = \sum_{i=1}^n  \ell( y_i , f_{\theta} (x_i) ) $, now the objective function cannot be decomposed. 
In this 1-dimensional example, there is only one feature at each layer.
For a high-dimensional neural network,
the BN operation applies to each feature of the pre-activations separately,
and aggregates information across samples to compute the mean and variance. 

\textbf{Second trick of BatchNorm}.  
In practice, the network cannot take all samples to compute the mean and variance,
thus it is natural to take samples in one mini-batch (say, 64 samples) to compute
the mean and variance. 
Suppose there are  $N$ samples in a mini-batch,
then the new network takes $N$ inputs jointly and produce $N$ predictions, i.e., $ ( \hat{y}_1, \dots , \hat{y}_N  ) =  \tilde{f}_{ \tilde{ \theta} }( x_1, \dots, x_N )  $.
Now the objective function can be decomposed across mini-batches. 
Mini-batch stochastic gradient descent (or other methods) can still be applied to the new objective function to learn the weights \footnote{See also Section 8.7.1 of \cite{goodfellow2016deep} for a description. }. 

\textbf{Inference stage}. 
Finally, there is a small trick for the inference stage. 
After training the network, one needs to perform inference for new data (e.g. predict the labels
of unseen images).  If we rigorously follow the paradigm of training/test, 
then we need to take a mini-batch of test samples as input to the network.
Nevertheless, in practice one often uses the mean and variance computed for the training data,
and thus the network can make prediction for each single test sample.

\section{Theoretical Complexity of Large-scale Optimization Methods}\label{appen: review of large scale optimization}
In this section, we review the theoretical complexity of a few optimization
methods for large-scale optimization. 
We explain the explicit convergence rate and computational complexity,
in order to reveal the connection and differences of various methods. 

To unify these methods in one framework, we start with the common
 convergence rate results of gradient descent method (GD) and explain how different methods improve the convergence rate in different ways. 
Consider the prototype convergence rate result in convex optimization: the epoch-complexity \footnote{For batch GD, one epoch is one iteration. For SGD, one epoch consists of multiple stochastic gradient steps that pass all data points once.  We do not say``iteration complexity'' or ``the number of iterations'' since per-iteration cost for the vanilla gradient descent and SGD are different and can easily cause confusion. In contrast, the per-epoch cost (number of operations) for batch GD and SGD are comparable.} is $O(  \kappa \log 1/\epsilon )$ or $O( \beta /\epsilon)$. These rates mean the following: to achieve $\epsilon$ error, the number of epochs
to achieve error $\epsilon$ is no more than $ \kappa \log 1/\epsilon $ for strongly convex problems (or $\beta/\epsilon$ for convex problems), where $\kappa $ is the
condition number of the problem (and $\beta$ is the global Lipschitz constant of the gradient).
For simplicity, we focus on strongly convex problems.

There are at least four classes of methods that can improve the computation time of GD
for strongly convex problems \footnote{Note that the methods discussed below also improve the rate for convex problems but we skip the discussions on convex problems.}.

The first class of methods are parallel computation methods.
This method mainly saves the per-epoch computation time, instead of improving
the overall convergence speed. 
For example,  for minimizing an $n$-dimensional least square problem, 
each epoch of GD requires a matrix-vector product which is parallelizable.
 More specifically, while a serial implementation takes time $O(n^2)$ to perform a matrix-vector product, a parallel model can take time as little as $O(\log n )$. 
This is a simplified discussion, and many other factors such as 
the computation graph of the hardware, synchronization cost and the communication cost can greatly affect the performance.
In general, parallel computation is quite complicated, which is why  an area called parallel computation is devoted to this topic (see the classical book \cite{bertsekas1989parallel} for
an excellent discussion of the intersection of parallel computation and numerical optimization). 
For deep learning, as discussed earlier, using $K$ machines to achieve nearly $K$-times
 speedup has been a popular thread of research.

The second class of methods are fast gradient methods (FGM) that have convergence rate $O( \sqrt{ \kappa} \log 1/\epsilon )$, thus saving a factor of $\sqrt{\kappa}$ compared to the convergence rate of GD $O(  \kappa \log 1/\epsilon )$. FGM includes conjugate gradient method, heavy ball method and accelerated gradient method.  
For convex quadratic problems, these three methods all achieve the improved rate $O( \sqrt{ \kappa} \log 1/\epsilon )$.
For general strongly convex problems, only accelerated gradient method  is known to achieve the rate $O( \sqrt{ \kappa} \log 1/\epsilon )$. 

The third class of methods are based on decomposition, i.e. decomposing a large problem into smaller ones. 
Due to the hardware limit and the huge number of data/parameters, 
it is often impossible to  process all samples/parameters at once,
thus loading data/parameters separately becomes a necessity. 
In this serial computation model,  GD can still be implemented (e.g. by gradient accumulation),
but it is not the fastest method. 
Better methods are 
 decomposition-based methods, including SGD, coordinate descent (CD)
and their mixture. 
To illustrate their theoretical benefits, consider an unconstrained $ d$-dimensional least squares problem with $n  $ samples.
For simplicity, assume $ n  \geq  d$ and the Hessian matrix has rank $ d $. 

\begin{itemize}
	\item  Randomized CD has an epoch-complexity $O(\kappa_{\mathrm D} \log 1/\epsilon )$ \cite{leventhal2010randomized,nestrov12}, where $\kappa_{\mathrm D}$ is the ratio of the average eigenvalue $\lambda_{\text{avg}}$ over the minimum eigenvalue $\lambda_{\min }$ of the coefficient matrix, and is related to Demmel's condition number. This is smaller than the rate of GD $O(\kappa \log 1/\epsilon )$ by a factor of
	$ \lambda_{\max}/\lambda_{\text{avg}} $ where $\lambda_{\max} $ is the maximum eigenvalue. Clearly, the improvement ratio $ \lambda_{\max}/\lambda_{\text{avg}} $ lies
	in $[1,   d]$, thus randomized CD is $ 1 $  to $ d $ times faster than GD. 
	
	\item  
	Recent variants of SGD (such as SVRG \cite{johnson2013accelerating} and SAGA \cite{defazio2014saga}) achieve an epoch-complexity 
	   $O( \frac{n}{d}  \kappa_{\mathrm D} \log 1/\epsilon )$, 
	  which is 1 to $n$ times faster than GD.
  	When $n = d$, this complexity is the same as R-CD for least squares
	problems (though not pointed out in the literature).
	We highlight that this up-to-$n$-factor acceleration has been the major focus of recent studies of SGD type methods, and has achieved much attention in theoretical machine learning area. 
	
	\item Classical theory of vanilla SGD \cite{bottou2018optimization} often uses diminishing stepsize and thus does not enjoy the same benefit as SVRG and SAGA.
	However, as discussed earlier,  for realizable least squares problem,
	 SGD with constant step-size can achieve an epoch-complexity
	    $O( \frac{n}{d}  \kappa_{\mathrm D} \log 1/\epsilon )$,  which is 
	    1 to $n$ times faster than GD. 
\end{itemize}

The above discussions are mainly for single sample/coordinate algorithms.
If the samples/coordinates are grouped in mini-batches/blocks,
 the maximal acceleration ratio is roughly the number of mini-batches/blocks.

The fourth class of methods utilize the second order information of the problem, including quasi-Newton method and  GD with adaptive learning rates. 
Quasi-Newton methods such as BFGS and limited BFGS (see, e.g., \cite{wright1999numerical}) use an approximation of the Hessian in each epoch, and are popular choices for many nonlinear optimization problems. 
AdaGrad, RMSProp, Adam and Barzilai-Borwein (BB) method use a diagonal  matrix estimation
 of the Hessian. 
 It seems very difficult to theoretically justify the advantage of these methods over GD, but intuitively, the convergence speed of these methods rely much less on the condition number $\kappa $ (or any variant of the condition number such as $\kappa_{\mathrm D}$). 



We briefly summarize the benefits of  these methods as below. Consider minimizing $d$-dimensional least squares problem with $n = d$ samples, and suppose
each machine can process at most one sample or one coordinate at once, 
which takes one time unit. 
The benchmark GD takes time $O( n \kappa \log 1/\epsilon )$ to achieve error $\epsilon $. 
With $n$ machines,  the first class  (parallel computation) can potentially reduce the computation time to $O(  \kappa \log n \log 1/\epsilon )$ with extra cost such as communication. 
For other methods, we assume only one machine is available. 
The second class (e.g. accelerated gradient method) reduces the computation time
 to $O( n \sqrt{ \kappa } \log 1/\epsilon )$, the third  class (e.g. SVRG and R-CD) 
 reduces the computation time to $O( n \kappa_{\mathrm D} \log 1/\epsilon )$, and the fourth class (e.g. BFGS and BB) may improve $\kappa$ to other parameters that are unclear. 
Although we treat these methods as separate classes, researchers have extensively studied various mixed methods of two or more classes, though the theoretical analysis can be much harder.

\newpage

\small
\bibliographystyle{plainnat}

\bibliography{biblio,ref}

\end{document}